\documentclass[twoside]{article}

\usepackage[accepted]{aistats2024}
\usepackage{tabulary}
\usepackage{hyperref}
\usepackage{url}
\usepackage{mathrsfs} 
\usepackage{ifthen}
\usepackage{graphicx}
\usepackage{epsfig}
\usepackage{dsfont, amssymb, mathtools}
\usepackage{amsfonts,dsfont,amssymb,amsthm,stmaryrd,bbm}
\usepackage{amsmath}
\usepackage{natbib}
\usepackage{subcaption}
\usepackage{xcolor}
\usepackage{comment}
\usepackage{algorithm,algorithmic}
\usepackage{wrapfig}

\newtheorem{thm}{Theorem}[section]

\newtheorem{remark}[thm]{Remark}
\newtheorem{assumption}[thm]{Assumption}
\newtheorem{lem}[thm]{Lemma}

\newtheorem{defn}[thm]{Definition}
\newtheorem{cor}[thm]{Corollary}

\newcommand{\cH}{\mathcal{H}} 
\newcommand{\mP}{\mathbb{P}}

\newcommand{\mC}{\mathbb{C}}

\newcommand{\mbR}{\mathbb{R}}

\newcommand{\mbZ}{\mathbb{Z}}
\newcommand{\mbS}{\mathbb{S}}
\newcommand{\mbE}{\mathbb{E}}

\newcommand{\cC}{\mathcal{C}}

\newcommand{\cF}{\mathcal{F}}
\newcommand{\cS}{\mathcal{S}}
\newcommand{\cG}{\mathcal{G}}
\newcommand{\1}{\mathbb{I}}

\newcommand{\calN}{\mathcal{N}}

\newcommand{\A}{\mathbf{A}}
\newcommand{\B}{\mathbf{B}}
\def\e{{\bf e}}

\def\Q{{\bf Q}}
\def\H{{\bf H}}
\def\I{{\bf I}}
\def\T{{\bf T}}
\def\S{{\bf S}}
\def\L{{\bf L}}
\def\w{{\bf w}}

\def\x{{\bf x}}
\def\y{{\bf y}}
\def\brx{\breve{\x}}
\def\bre{\breve{\e}}
\def\U{{\bf U}}

\def\phi{\varphi}

\newcommand\hatt[1]{\widehat{#1}} 

\begin{document}

\twocolumn[

\aistatstitle{Information Theoretically Optimal Sample Complexity of Learning \emph{Dynamical} Directed Acyclic Graphs}

\aistatsauthor{ Mishfad Shaikh Veedu \And Deepjyoti Deka \And Murti V Salapaka }

\aistatsaddress{University of Minnesota, Twin Cities \And Los Alamos National Laboratory \And University of Minnesota, Twin Cities} ]

\begin{abstract}
 In this article, the optimal sample complexity of learning the underlying interactions or dependencies of a Linear Dynamical System (LDS) over a Directed Acyclic Graph (DAG) is studied. We call such a DAG underlying an LDS as \emph{dynamical} DAG (DDAG). In particular, we consider a DDAG where the nodal dynamics are driven by unobserved exogenous noise sources that are wide-sense stationary (WSS) in time but are mutually uncorrelated, and have the same {power spectral density (PSD)}. Inspired by the static DAG setting, a metric and an algorithm based on the PSD matrix of the observed time series are proposed to reconstruct the DDAG. It is shown that the optimal sample complexity (or length of state trajectory) needed to learn the DDAG is $n=\Theta(q\log(p/q))$, where $p$ is the number of nodes and $q$ is the maximum number of parents per node. To prove the sample complexity upper bound, a concentration bound for the PSD estimation is derived, under two different sampling strategies. A matching min-max lower bound using generalized Fano's inequality also is provided, thus showing the order optimality of the proposed algorithm. The codes used in the paper are available at {\href{https://github.com/Mishfad/Learning-Dynamical-DAGs}{https://github.com/Mishfad/Learning-Dynamical-DAGs}}
\end{abstract}

\section{Introduction}
{Learning the interdependency structure in a network of agents}, from passive time series observations, is a salient problem with applications in neuroscience \cite{bower2012book}, finance \cite{kim2011time}, meteorology \cite{climate}, etc. Reconstructing the exact structure with the dependency/causation directions has a wide range of applications. For example, the identification of causation structure among the shares helps in obtaining robust portfolio management in the stock market \cite{kim2011time}. Similarly, causal graphs are useful in understanding dynamics and identifying contributing factors of a public epidemic emergency situation \cite{yang2020transportation}.

The structure of directed interactions in a network of agents is conveniently represented using directed graphs, with the agents as nodes and the directed interactions as directed edges. If the underlying graph doesn't have cycles, it is called a Directed Acyclic Graph (DAG). In general, it is not possible to reconstruct the exact structure with the direction of different edges. Instead, in many networks it is possible to retrieve only the Markov equivalence graphs, the set of graphs satisfying the same conditional dependence property, from data without any intervention \cite{ghoshal18a}. In applications such as finance \cite{kim2011time}, climate science \cite{climate} etc., the agent states, instead of being temporally independent, can evolve over time due to past directed interactions. Such temporal evolution can be represented by a linear dynamical system (LDS). In LDS, the interaction between agent states is captured by a linear time-invariant function. In this paper, we study the identifiability and present the first sample complexity results for learning a DAG of LDS, which we term as \emph{Dynamical} DAG or \textbf{DDAG}. This is distinguished from \emph{static} DAG, where the agent states are temporally independent and the DAG does not correspond to temporal dynamics.

\subsection{Related Work} 
\textbf{Static DAG Learning:} The problem of obtaining an upper bound on the sample complexity of learning static DAGs goes back twenty-five years \cite{friedman1996sample}, \cite{zuk2006number}. However, tight characterization of optimal rates for DAG learning is a harder problem compared to undirected networks \cite{gao2022optimal}, primarily due to the order identification step. Identifiability conditions for learning static DAGs with linear interactions and excited by equal variance Gaussian noise were given in \cite{peters2014identifiability}. Several polynomial time algorithms have been proposed for static DAG reconstruction using samples of states at the graph nodes; see \cite{ghoshal2017information,ghoshal2017learning,ghoshal18a,chen2019causal,gao2022optimal,park2020identifiability,park2017learning}, \cite{runge2019inferring} \cite{sugihara2012detecting} \cite{shimizu2006linear} \cite{spirtes2016causal}, and the reference therein. An information-theoretic lower bound on structure estimation was studied in \cite{ghoshal2017information}. In \cite{gao2022optimal}, it was shown that the order optimal sample complexity for static Gaussian graphical model with equal variance is $n=\Theta(q\log (p/q))$, where $p$ is the number of nodes and $q$ is the maximum number of parents. The authors showed that the algorithm given in \cite{chen2019causal} provides an upper bound that matches a min-max lower bound for the number of samples. However, similar results for DDAGs, with underlying temporal directed interaction between agent states, have not been studied, to the best of our knowledge. 

\textbf{LDS Learning:} Graph reconstruction, in general, is challenging in a network of LDS (undirected, or bi-directed), as it involves time-dependencies between collected samples of nodal states. Learning the conditional independence structure in LDS with independent and identically distributed (white) excitation was explored in \cite{basu2015regularized}, \cite{loh2011high}, \cite{songsiri2010topology}, \cite{simchowitz2018learning}, \cite{faradonbeh2018finite}, and the references therein. However, the methods in the cited papers do not extend to LDS that is excited by WSS (wide-sense stationary) noise, which makes correlations in state samples more pronounced. For LDS with WSS noise, \cite{tank2015bayesian}, \cite{dahlhaus2000graphical}, \cite{materassi2013reconstruction}, and \cite{MATERASSI_innocenti_physica}, estimated the conditional correlation structure, which contains true edges in the network and extra edges between some two-hop neighbors \cite{materassi_tac12}. A consistent algorithm for the recovery of exact topology in LDS with WSS noise was provided in \cite{TALUKDAR_physics}, with the corresponding sample-complexity analysis developed in \cite{doddi2022efficient}, using a neighborhood-based regression framework. However, the developed algorithms and related sample complexity results do not extend to directed graphs and hence exclude DDAG reconstruction. DDAG reconstruction from the time-series data has been explored using the framework of directed mutual information in \cite{Quinn_DIG} but without rate characterization for learning from finite samples. Recently, extending the results in \cite{chen2019causal}, \cite{dallakyan2023learning} studied the graph reconstruction for Discrete Time Fourier Transform (DTFT) models, a special case of LDS with i.i.d. noise, but 
again without rate characterization for learning from finite samples. 

\textbf{Contribution:} This article presents an information-theoretically optimal sample complexity analysis for learning a dynamical DAG (DDAG) excited by wide-sense stationary (WSS, i.e., non i.i.d) noise of equal power spectral density, using samples of state trajectories of the underlying linear dynamical system (LDS). To the best of our knowledge, this is the first paper to study and prove sample complexity analysis for DDAGs. We consider learning under two sampling scenarios, viz; 1) restart and record, 2) continuous sampling. While the former pertains to samples collected from disjoint (independent) trajectories of state evolution, the latter includes samples from a single but longer state trajectory (see Fig.~\ref{fig:plots}). Surprisingly, the results in this article show that the estimation errors are not influenced by the sampling strategy (restart and record, or continuous) as long as the number of collected samples are over a determined threshold given by $n=O(q\log(p/q))$, where $p$ is the number of nodes and $q$ is the maximum number of parents per node. We also provide a matching information-theoretic lower-bound, $\max\left(\frac{ \log p}{2\beta^2+ \beta^4},\frac{ q\log (p/q)}{M^2-1}\right)$, where $\beta$ and $M$ are system parameters (see Definition \ref{def:H_p,q}); thus
obtaining an order optimal bound $n=\Theta(q\log(p/q))$. 

Our learning algorithm relies on first deriving rules for DDAG estimation using the Power Spectral Density Matrix (PSDM) of nodal states, inspired by the estimator for static DAGs based on covariance matrices \cite{chen2019causal}. Subsequently, the sample complexity associated with learning is derived by obtaining concentration bounds for the PSDM. In this regard, characterization of non-asymptotic bounds of PSDMs for a few spectral estimators have been obtained in \cite{fiecas2019spectral,doddi2019topology,zhang2021convergence} previously. A unified framework of concentration bounds for a general class of PSDM estimators was recently presented in \cite{lamperski2023non}. Our concentration bounds of the PSDM are reached using different proof steps, based on Rademacher random variables and symmetrization argument \cite{wainwright_2019}.

The rest of the paper is organized as follows. Section 2 introduces the system model and the preliminary definitions for LDS and DDAGs. Section 3 discusses Algorithm \ref{alg:Ordering} and the main results for DDAG reconstruction from PSDM. Section 4 provides a concentration bound for the error in estimating the PSDM and a sample complexity upper bound for DDAG reconstruction using Algorithm \ref{alg:Ordering}. Section 5 contains a sample complexity lower bound.

\emph{Notations:} Bold faced small letters, $\x$ denote vectors; Bold faced capital letters, $\A$ denote matrices; For a time-series, $\x$, $\breve{x}(t)$ denotes the value of $\x$ at time $t$, $\x(\omega)$ denotes the discrete time Fourier transform of $\x$, $\x(\omega):=\sum_{k=-\infty}^\infty \breve{\x}(k) e^{-i\omega k}$, $\omega \in \Omega=[0,2\pi]$; $diag(v_1,\dots,v_p)$ operator creates a diagonal matrix with diagonal entries $v_1,\dots,v_p$; $\Phi_{\mathbf{x}_{AB}}$ or $\Phi_{{AB}}$ denotes the matrix obtained by selecting rows $A$ and columns $B$ in $\Phi_\x$; $\A^*$ denotes conjugate transpose of $\A$. $\|\A\|$ is the spectral norm of $\A$ and $\|\mathbf{v}\|_2$ is the Euclidean norm of vector $\mathbf{v}$.

\section{System model of DDAG}
We describe different aspects of DDAG (DAG underlying an LDS) and the sampling strategies considered. We begin with some necessary DAG terminologies. 

\begin{figure}[htb!]
\centering
\includegraphics[trim=70 140 680 335,clip, width=.7\linewidth]{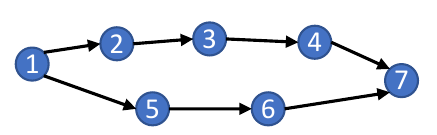} \caption{An example DDAG. Node 1 is an ancestor and node 7 is a descendant of every node in the graph. The set $\{1,2,5\}$ is an ancestral set but $\{2,5\}$ is not. $an(3)=\{1,2\}$, $desc(3)=\{4,7\}$, $nd(3)=\{1,2,5,6\}$.}
\label{fig:dag} \end{figure}

\textbf{DDAG terminology:} The \emph{dynamical} directed acyclic graph (DDAG), is given by $G:=(V,E)$, where node set $V=\{1,\dots,p\}$, and $E$ is edge set of directed edge $i \xrightarrow{} j$. A directed path from $i$ to $j$ is a path of the form $i=v_0\xrightarrow{} v_1\xrightarrow{} \dots \xrightarrow{} v_\ell \xrightarrow[]{}v_{\ell+1}=j$, where $v_k \in V$ and $(v_k,v_{k+1}) \in E$ for every $k=0,\dots,\ell$. A cycle is a directed path from $i$ to $i$, which does not exist in DDAG $G$. For $G$, $pa(i):=\{j \in V: (j,i) \in E\}$ denotes the parents set and $desc(i)$ denotes the descendants of $i$, the nodes that have a directed path from $i$. The set $nd(i):=V \setminus desc(i)$ denotes the non-descendants set and the set $an(i)\subset nd(i)$ denotes the ancestors set, the nodes that have a directed path to $i$. At set $C \subseteq V$ is called ancestral if for every $i \in C$, $pa(i) \subseteq C$. {Figure \ref{fig:dag} shows an example DDAG with the node definitions.} 

An ordered node set $C \subseteq V$ is said to be a topological ordering on $G$ if for every $i,j \in C$, $i\in desc(j)$ in $G$ implies $i>j$. $\cG_{p,q}$ denotes the family of DDAGs with $p$ nodes and at most $q$ parents per node.
Without a loss of generalizability, we use the same terminology for the DDAG and the underlying DAG.

\textbf{LDS Model excited by equal PSD WSS noise:} For the DDAG $G=(V,E) \in \cG_{p,q}$, we consider a linear dynamical system (LDS) with $p$ scalar state variables, corresponding to nodes in $V$. Node $i$'s states correspond to $\{\breve{x}_i(k)\}_{k \in \mbZ},~ 1 \leq i\leq p$. The LDS evolves according to the linear time-invariant model,
\begin{align}
\label{eq:convolution_model}
 \breve{x}_i(k)=\sum_{(j,i) \in E, j \neq i}^p (\breve{h}_{ij} \star \breve{x}_j)(k)+\breve{e}_i(k), ~k\in \mbZ,
\end{align}
where transfer function $\breve{h}_{ij}\neq 0$ when directed edge $(j,i) \in E$. The exogenous noise $\{\breve{e}_i(k)\}_{k \in \mbZ},~ 1 \leq i\leq p,$ are zero mean wide sense stationary Gaussian processes, uncorrelated across nodes. Taking the discrete-time Fourier transform (DTFT) of \eqref{eq:convolution_model} provides the frequency representation for every $\omega \in \Omega=[0,2\pi]$,
\begin{align}
\label{eq:LDS_i}
 \x_i(\omega)=\sum_{(j,i) \in E, j \neq i}^p \H_{ij}(\omega) \x_j(\omega )+\e_i(\omega), ~ 1\leq i\leq p,
\end{align}
where $\x_i(\omega)=\cF\{\breve{x}_i\}:=\sum_{k=-\infty}^\infty \breve{x}_i(k) e^{-i\omega k}$, $\e_i(\omega)=\cF\{\breve{e}_i\}$, and $\H_{ij}(\omega)=\cF\{\breve{h}_{ij}\}$.
The model in \eqref{eq:LDS_i} can be represented in the matrix form to obtain the following LDS,
\begin{align}
\label{eq:LDS}
 \mathbf{x}(\omega)=\mathbf{H}(\omega)\mathbf{x}(\omega)+\mathbf{e}(\omega), ~ \forall \omega \in \Omega,
\end{align}
where $\mathbf{e}(\omega)$ is the WSS noise. In this article, we are interested in the LDS with $\Phi_{\mathbf{e}}(\omega)=\sigma(\omega) diag(\alpha_1,\dots,\alpha_p)$, where $\alpha_i$ are known and can be a function of $\omega$. For the simplicity of analysis, henceforth it is assumed that $\Phi_{\mathbf{e}}(\omega)=\sigma(\omega)\I$. 
\begin{remark}
 The assumption $\Phi_\e(\omega) = \sigma(\omega)\I$ is a restrictive assumption. However, we would like to remark that some form of restriction is required for DAG reconstruction in a general setup due to identifiability issues \cite{shimizu2006linear}. The assumption can be relaxed using ordered conditions on $\Phi_{e_i}$ and $\H$, similar to \cite{ghoshal18a}. Furthermore, our results on DDAG reconstruction require the equal PSD to hold \emph{only at some known $\omega \in \Omega$}, which is less restrictive.
\end{remark}

The power spectral density matrix (PSDM) of the time-series $\x$ at the angular frequency $\omega \in \Omega$ is given by \begin{align}
\Phi_{\mathbf{x}}(\omega)=\mathcal{F}\left\{R_\x(t)\right\}=
\sum_{k=-\infty}^\infty R_\x(k) e^{-i\omega k},
\end{align}
where $R_\x(k):=\mbE[\breve{\x}(k)\breve{\x}^T(0)]$ is the auto-correlation matrix of the time-series $\x$ at lag $k$. The $(i,j)$-th entry of $\Phi_\x$ is denoted by $\Phi_{ij}$. For the LDS \eqref{eq:LDS}, the PSDM is given by
\begin{align}
 \Phi_{\mathbf{x}}(\omega)=(\I-\mathbf{H}(\omega))^{-1} \Phi_{\mathbf{e}}(\omega) ((\I-\mathbf{H}(\omega))^{-1})^*.
\end{align}
Consider the following non-restrictive assumptions on the power spectral density and correlation matrix of the LDS states. 
\begin{assumption}
\label{ass:psdm}
There exists a $M \in \mbR$ such that $\frac{1}{M} \leq \lambda_{min}({\Phi}_\x)\leq \lambda_{max}({\Phi}_\x)\leq M$, where $\lambda_{min}$ and $\lambda_{max}$ respectively denote the minimum and maximum eigenvalues. 
\end{assumption}

\begin{assumption}
\label{ass:autocorrelation}
The auto-correlation matrix of the time-series $\x$ at lag $k$, 
$R_\x(k):=\mbE[\breve{\x}(k)\breve{\x}^T(0)]$ satisfies $\|R_\x(k)\| \leq C \rho^{-|k|}$, for some positive constants $C,\rho\in \mbR$, $\rho>1$. \end{assumption}
In the remaining paper, following these assumptions, our interest will be limited to the following family of DDAGs and corresponding LDS.
\begin{defn}
\label{def:H_p,q} 
$\cH_{p,q}(\beta,\sigma,M)$ denotes the family of LDS given by \eqref{eq:LDS} such that the corresponding DDAG, $G(V,E) \in \cG_{p,q}$ ($p$ nodes with each node having a maximum $q$ parents), with $|\H_{ij}(\omega)| \geq \beta, ~\forall (j,i)\in E$, $\Phi_\e(\omega) = \sigma(\omega)\I$, and $M^{-1} \leq \lambda_{\min}(\Phi_\x(\omega)) \leq \lambda_{max}(\Phi_\x(\omega)) \leq M$, $\forall \omega \in \Omega$
\end{defn}

\begin{figure}[htb!]
\centering
\includegraphics[trim=190 290 180 200,clip, width=\linewidth]{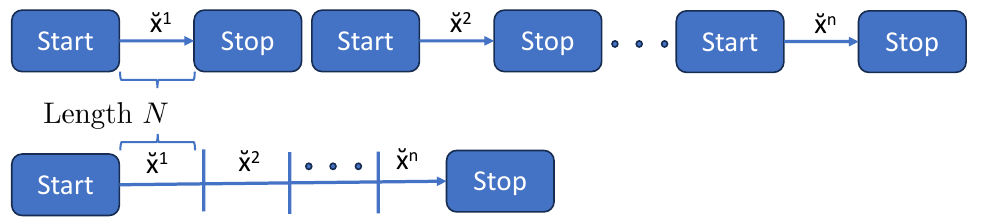} \caption{(a) shows restart and record sampling. (b) shows continuous sampling.}
\label{fig:sampling}
\end{figure}

\textbf{Sampling Strategy for LDS states:} We consider two sampling settings (see Fig.~\ref{fig:sampling} for details):\\
i) \textbf{restart and record:} The sampling is performed as follows: start recording and stop it after $N$ measurements. For the next trajectory, the procedure is restarted with an independent realization and record for another epoch of $N$ samples; repeat the process another $n-2$ times providing $n$ i.i.d trajectories of $N$ samples each. \\
ii) \textbf{{continuous sampling:}} Here, a single trajectory of length $n \times N$ is taken consecutively. Then, the observations are divided into $n$ segments, each having $N$ consecutive samples.

The data collected using either strategy is thus grouped into $n$ trajectories of $N$ samples each. For the finite length $r^{\text{th}}$ trajectory, $\{\breve{x}^r(t)\}_{t=0}^{N-1}$, define for each $\omega \in \Omega$,
\begin{align}
\label{eq:def_x^r(omega)}
 \x^r(\omega)=\frac{1}{\sqrt{N}} \sum_{k=0}^{N-1} \breve{x}^r(k) e^{-i\omega k}, r=1,\dots, n. 
\end{align}
Note that $\x^r(\omega)$, if exists, is a zero mean random variable with the covariance matrix given by 
\begin{align}
 \widetilde{\Phi}_{\x}(\omega)&:=\mbE\left\{ \x^r(\omega) [\x^r(\omega)]^* \right\}
\label{eq:DFT_covariance}\\
 &= \frac{1}{N} \sum_{k=-(N-1)}^{N-1} (N-|k|) R_\x(k) e^{-i\omega k}, \forall \omega \in \Omega.\nonumber
\end{align}
For the restart and record setting (unlike for continuous sampling), $\{\x^r(\omega)\}_{r=1}^n$ are i.i.d.
Further, as $N \xrightarrow{} \infty$, $\widetilde{\Phi}_{\x}(\omega) \xrightarrow{} {\Phi}_{\x}(\omega)$ uniformly in $\Omega$ \cite{stoica2005spectral}.

\section{Reconstructing DDAGs from PSDM}
In this section, we discuss some results on how the PSDM, $\Phi_\x$, can be employed to completely reconstruct the DDAG $G$, when the time series is generated according to \eqref{eq:LDS}. Applying these results, inspired from the algorithm for static setting \cite{chen2019causal}, we propose Algorithm \ref{alg:Ordering} for reconstructing the DDAG. First, we prove that conditional PSD (CPSD) of $i$ conditioned on $C$, defined as 
\begin{align}
 \label{eq:f}
 f(i,C,\omega):=\Phi_{ii}(\omega)- \Phi_{iC}(\omega) \Phi_{CC}^{-1}(\omega)\Phi_{Ci}(\omega),
\end{align}
is a metric sufficient to obtain a topological ordering on the DDAG $G$, which aids in the reconstruction of $G$. Notice that unlike \cite{chen2019causal}'s static setting that uses conditional covariance matrices, our algorithm uses CPSD to unveil the dependencies in the DDAG. This further affects the sample complexity analysis, as discussed subsequently.
\subsection{CPSD and Topological Ordering}
Here, we will show that CPSD is the minimum for the nodes that have all the parents included in the conditioning set. We start by proving the result for the source nodes.
\begin{lem}
\label{lem:PSD_min}
Consider the LDS described by \eqref{eq:LDS}. For any $\omega \in \Omega$, let $\alpha^*:=\min\limits_{k\in V}\Phi_{kk}(\omega)$. Then $\Phi_{ii}(\omega)=\alpha^*$ if and only if $i$ is a source node.
\end{lem}
\proof
Let $\T(\omega)=(\I-\H(\omega))^{-1}$ and $\Phi_{\e}(\omega)=\sigma(\omega) \I_n$. Then, $\Phi_{\mathbf{x}}(\omega)= \sigma(\omega) \T(\omega)\T^*(\omega)$. By Cayley-Hamilton theorem, there exists constants $a_0(\omega),\dots,a_n(\omega)$ such that 
$ (\I-\H(\omega))^{-1}=\sum_{k=0}^{n-1} a_k(\omega) (\I-\H(\omega))^k$. Using induction, it can be shown that non-diagonal entries of $(\I-\H(\omega))^k$ are zero if and only if there no $k-hop$ path between $i$ and $j$ (almost always) between $j$ and $i$. Similarly, $(i,i)$th entry is $1$ if and only if there is no k-hop path between them (almost always).

Then (ignoring $(\omega)$), $\Phi_{ii}=(\Phi_{\mathbf{x}})_{ii}=\sigma \sum_{k=1}^n \T_{ik}\T_{ik} =\sigma(\T_{ii}^2+\sum_{k\neq i} \T_{ik}^2)$. If $i$ is a source node, then $\T_{ik}=0$ for every $k\neq i$ and $\T_{ii}=1$, which implies $\Phi_{ii}=\sigma \T_{ii}^2$. For non-source nodes, $\T_{ik}\neq 0$ for some $k$, which gives $\Phi_{ii}> \sigma$. \hfill \qedsymbol

\subsubsection{Conditional PSD deficit}
\label{subsection:CPSD deficit}

The following is the definition of conditional PSD deficit, which is helpful in proving the subsequent results and in retrieving the DDAG. \begin{defn}[The CPSD deficit]
\begin{align}
\label{eq:Delta_defn}
 \Delta:=\min_{\omega \in [0,2\pi]} \min_{j \in V} \min_{\substack{C\subset nd(j),\\
Pa(j) \setminus C \neq \emptyset}} f(j,C,\omega)-\sigma(\omega)
\end{align}
\end{defn}
The following lemma shows that $f(i,C,\omega)$ from Eq.~\ref{eq:f} can be used as a metric to obtain a topological ordering of $G$.

\begin{lem}
\label{lem:f_comparison}
Consider the DDAG, $G$, governed by \eqref{eq:LDS}. Let $j \in V$ and let $C \subseteq V\setminus \{j\}$ be an ancestral set. Then for every $\omega \in \Omega$, 
\begin{align*}
 f(j,C,\omega)&=\sigma(\omega) &:\text{ if } Pa(j)\subseteq C,\\
 f(j,C,\omega)&\geq \Delta+\sigma(\omega)>\sigma(\omega) &:\text{ if } Pa(j)\nsubseteq C.
 \end{align*}
\end{lem}
The detailed proof is given in the supplementary material, Section 1. The proof uses the fact that conditioning on a superset of all parents makes node $j$ conditionally independent of other ancestors and leaves only the exogenous noise at the node as the only component of its CPSD.

\begin{cor}
\label{cor:f_diff}
If $Pa(i) \subseteq C$ and $Pa(i)\nsubseteq D$, then $f(i,D,\omega)-f(i,C,\omega)\geq \Delta$.
\end{cor}

\begin{lem}
\label{lem:Delta}
$\Delta \geq \beta^2\sigma$.
\end{lem}
\begin{proof}
The proof follows from Lemma \ref{lem:f_comparison}. Please see supplementary material for details.
\end{proof}

To determine the DDAG $G$'s structure, we first determine a topological ordering of nodes as follows: Beginning with an empty set $\cS$, we iteratively add node $i$ in $\cS$ where
\begin{align}
\label{eq:min_f}
(i,C_i^*)\in \mathop{\arg \min} \limits_{\substack{\ C \subseteq \cS, |C| \leq q\\ 1\leq j\leq p,\ j\notin \cS}} {f}(j,C,\omega), 
\end{align}
and $f$ comes from \eqref{eq:f}. The following Lemma shows that $\cS$ is a valid topological ordering w.r.t. $G$.


\begin{lem}
\label{lem:top_ordring_f}
 $\cS$ is a valid topological ordering with respect to $G$.
\end{lem}
\begin{proof}
 In the first step, $C=\emptyset$ and $f(i,C,\omega)=\Phi_{ii}$. By Lemma \ref{lem:PSD_min} and Lemma \ref{lem:f_comparison}, $\Phi_{ii}=\sigma$ if $i$ is a source node, where as $\Phi_{ii}\geq \sigma+\Delta$ if $i$ is not a source node. Thus, the first node in the ordered set $\cS$, $\cS_1$, is always a source node. Induction assumption: Nodes $\cS_1$ to $\cS_n$ in $\cS$ follow topological order. For the $n+1$, by Lemma \ref{lem:f_comparison}, for every $C \subseteq \cS$, $f(k,C,\omega)$ is minimum for $k \in V\setminus \cS$ if and only if $Pa(k) \subseteq C$. Thus nodes $\cS_1$ to $\cS_{n+1}$ follow a topological order, which proves the result.
\end{proof}

\noindent\underline{\textbf{Identification of the parents}}: Parents of a node are identified from the ordered set by applying Corollary \ref{cor:f_diff}. Let $D=C\setminus \{k\}$. As shown in Corollary \ref{cor:f_diff}, if $Pa(i) \subseteq C$ and $k \in Pa(i)$, then $f(i,C,\omega)-f(i,D,\omega)\geq \Delta$. Thus, from the set $\cS$, for every node $\cS_i$, one can eliminate nodes $\cS_1,\dots,\cS_{i-1}$ by checking if the difference is greater than $\Delta$. If the difference is greater than $\Delta$ for some $\cS_k$, then $\cS_k$ is a parent of $\cS_i$. That is,
\begin{lem}
\label{lem:parents}
Let $(i,C_i^*)$ be a solution of \eqref{eq:min_f} and let $$P_i:=\left\{ j \in C_i^* \left| ~ |f(i,C_i^*,\omega) - {f}(i,C_i^* \setminus j,\omega) | \geq \Delta \right. \right\}.$$ Then, $Pa(i)=P_i$.
\end{lem}

Applying the above procedure and Lemma \ref{lem:f_comparison}-Lemma \ref{lem:parents}, one can formulate Algorithm \ref{alg:Ordering} to obtain the ordering of the DDAG, $G$ and eventually reconstruct the DDAG exactly. In Algorithm \ref{alg:Ordering}, $\hatt{f}(i,C,\omega):=\hatt{\Phi}_{ii}(\omega)- \hatt{\Phi}_{iC}(\omega) \hatt{\Phi}_{CC}^{-1}(\omega)\hatt{\Phi}_{Ci}(\omega)$, an empirical estimate of $f(i,C,\omega)$, is employed instead of $f$ and $\gamma$ is taken as $\Delta/2$ (see Definition \eqref{eq:Delta_defn}). The following lemma proves that if the empirical estimate $\hatt{f}(\cdot)$ is close enough to the original $f(\cdot)$, then Algorithm \ref{alg:Ordering} reconstructs the DDAG exactly.

\begin{algorithm}
\caption{Reconstruction algorithm}
\noindent \hspace{0cm}\textbf{Input: } Estimated PSDM, $\widehat{\Phi}(\omega)$: $\Delta$, ~$\omega \in [0,2\pi]$\\
\hspace{-13cm}\textbf{Output:} $\hatt{G}$ 
\label{alg:Ordering}
\begin{enumerate}
\item Initialize the ordering, $\cS\xleftarrow{}$()
\item Repeat $p$ times
\begin{enumerate}
\item Compute $\left(j^*,C_j^*\right)\in \mathop{\arg \min} \limits_{\substack{\ C \subseteq \cS, |C| \leq q\\ 1\leq j\leq p,\ j\notin \cS}} \hatt{f}(j,C,\omega)$ 
\item $\cS \xleftarrow{} (\cS,~j^*)$
\end{enumerate}
\item $\hatt{G}=(V,\hatt{E}), V\xleftarrow{}\{1,\dots,p\}, \hatt{E}\xleftarrow{}\emptyset$
\item For $i=1,\dots,p$
\begin{enumerate}
\item {\footnotesize $P_i:=\left\{ j \in C_i^* \left| ~ |\hatt{f}(\cS_i,C_i^*,\omega) - \hatt{f}(\cS_i,C_i^* \setminus j,\omega) | \geq \gamma \right. \right\}$}
\item $\forall k\in P_i$, Do $\hatt{E}\xleftarrow{} \hatt{E} \cup \{(k,i)\}$
\end{enumerate}
\item Return $\hatt{G}$
\end{enumerate}
\end{algorithm}

\begin{lem}
\label{lem:G_reconstruction}
 If $|\hatt{f}(i,C,\omega)-{f}(i,C,\omega)| < \Delta/4$, for every $i,\omega,C$, then Algorithm \ref{alg:Ordering} reconstructs the DAG, $G$ successfully. That is, $G=\hatt{G}$.
\end{lem}
\begin{proof}
See supplementary material, section 2
\end{proof}
Therefore, it suffices to derive the conditions under which $|f(\cdot)-\hatt{f}(\cdot)|<\Delta/4$. In the following section, we derive a concentration bound to guarantee a small error, which in turn is applied in obtaining the upper bound on the sample complexity of estimating $G$.

\section{Finite Sample Analysis of Reconstructing DDAGs}

\label{sec:DFT_PSD_est}
In Lemma \ref{lem:G_reconstruction}, it was shown that the DDAG, $G$, can be reconstructed exactly if the error in estimating $f(i,C,\omega)$ (given by \eqref{eq:f}) is small enough. In this section, a concentration bound on the error in estimating $\Phi_\x$ from finite data is obtained, which is used later to obtain a concentration bound on the error in estimating the metric $f$. 

Recall that we consider $n$ state trajectories (see Fig.~\ref{fig:plots} for the two sampling strategies) with each trajectory being of length $N$ samples, i.e $\left\{\{\brx^r(k)\}_{k=0}^{N-1}\right\}_{r=1}^n$. For each trajectory, $\x^r(\omega)$, defined in \eqref{eq:def_x^r(omega)}, is a complex Gaussian with mean zero and covariance matrix, $\widetilde{\Phi}_x(\omega))$, i.e., for every $r=1,\dots,n$, $\x^r(\omega) \sim \mathcal{N}(0,\widetilde{\Phi}_\x(\omega))$, as given in Eq.~\ref{eq:DFT_covariance}. Increasing $N$ ensures that $\widetilde{\Phi}_\x$ is close to ${\Phi}_\x$. To estimate the PSDM, we thus rely on the spectrogram method and estimate $\widetilde{\Phi}_x(\omega))$ using finite $n$ samples for $\x^r$.

\subsection{Non-Asymptotic Estimation Error in Spectrogram Method}
Let $\hatt{\Phi}_x(\omega):=\frac{1}{n} \sum_{r=1}^n\x^r(\omega)[\x^r(\omega)]^*$. Let the estimation error in estimating $\Phi_\x$ by $\hatt{\Phi}_x(\omega)$ be $Q:= \hatt{\Phi}_x(\omega)-{\Phi}_x(\omega)$. 

Applying the triangle inequality, $\|Q\| \leq \|Q_{approx}\|+\|Q_{2}\|$, where $Q_{approx}:=\widetilde{\Phi}_x(\omega)-{\Phi}_x(\omega)$ and $Q_2:=\hatt{\Phi}_x(\omega)-\widetilde{\Phi}_x(\omega)$. Note that $\widetilde{\Phi}_x(\omega)$ is the covariance of $\x^r(\omega)$. To bound the estimation error, we bound both $Q_{approx}$ and $Q_2$. 

\begin{figure*}[htb!]
\begin{subfigure}{0.32\linewidth}
\includegraphics[trim=0 0 0 50,clip, width=\textwidth]{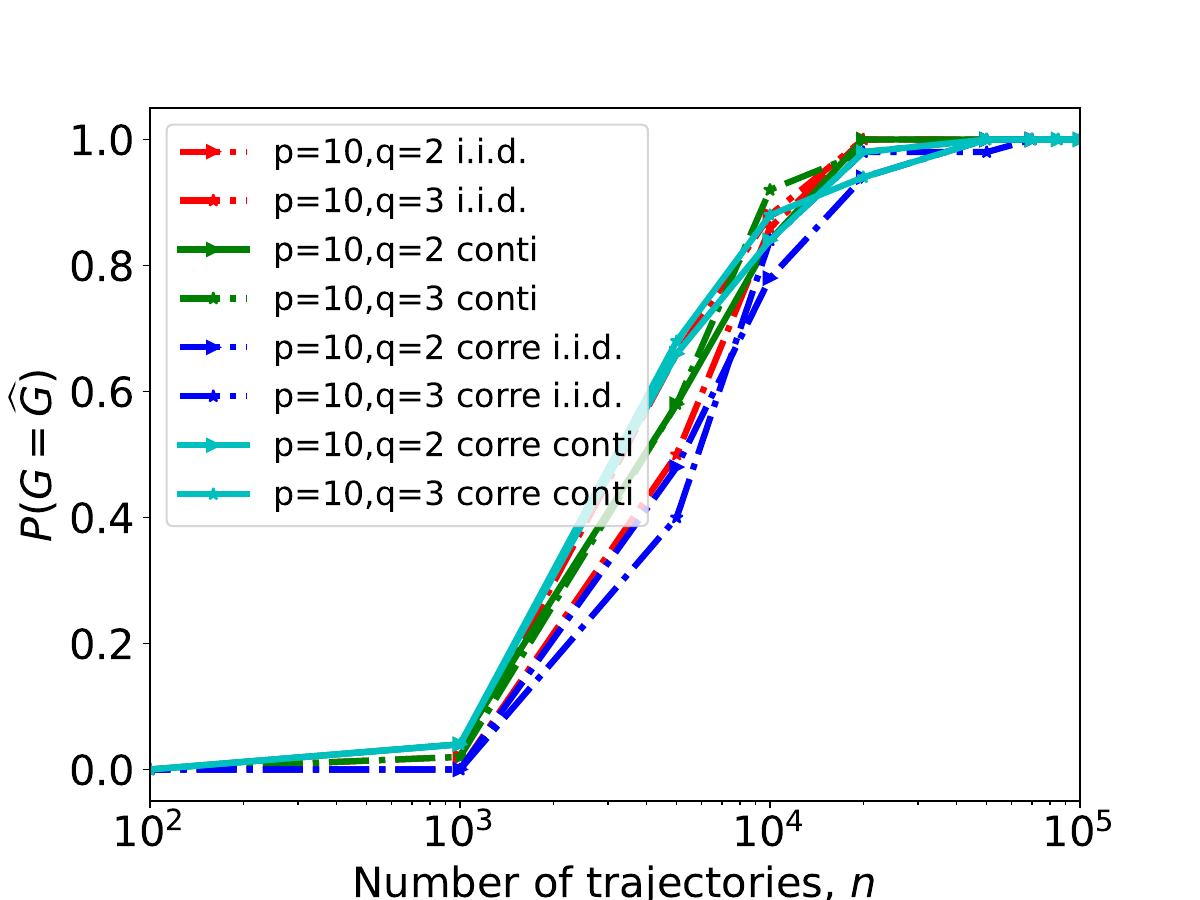} 
\caption{ $p=10$.}
\end{subfigure}
\hfill
\begin{subfigure}{0.32\linewidth}
\includegraphics[trim=0 0 0 50,clip, width=\textwidth]{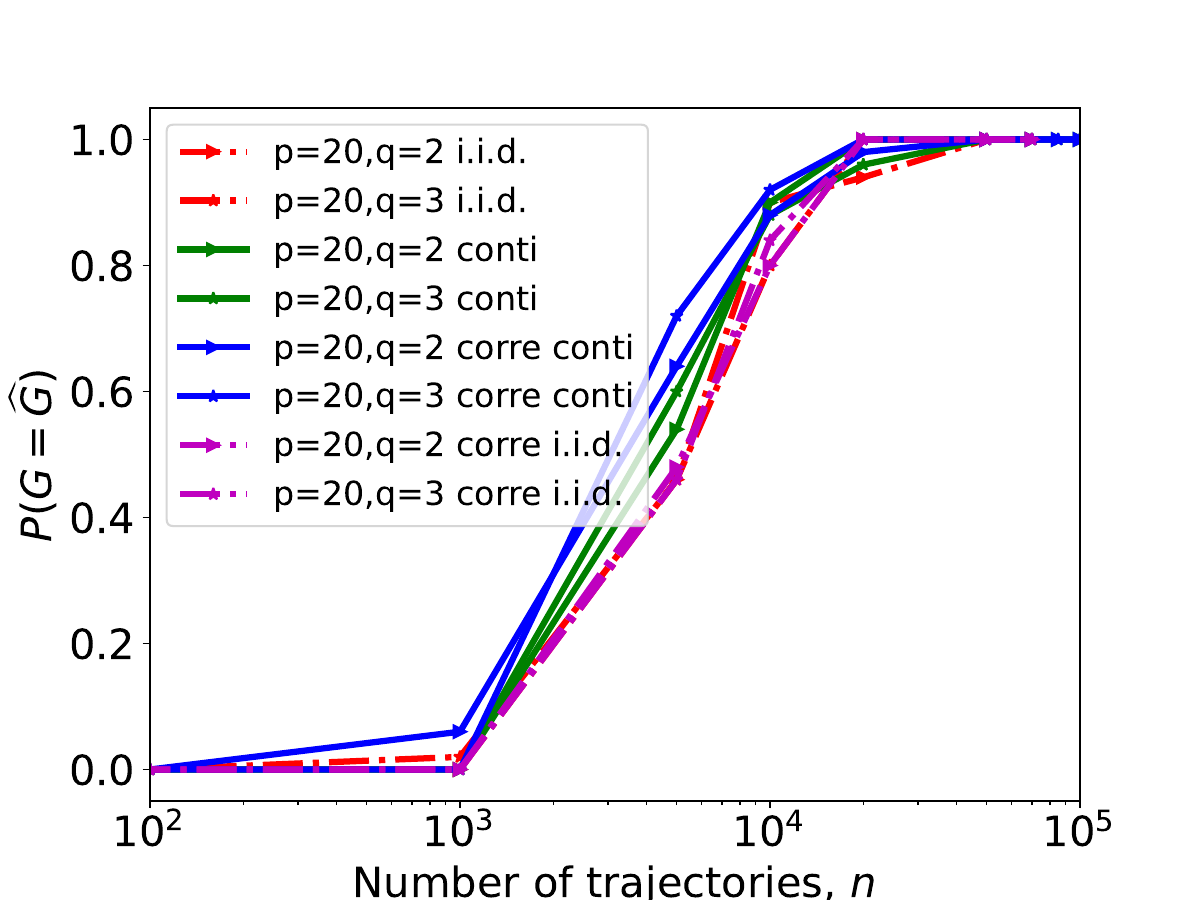} 
\caption{$p=20$.}
\end{subfigure}
\hfill
\begin{subfigure}{0.32\linewidth}
 \centering
\includegraphics[trim=0 0 0 50,clip, width=\textwidth]{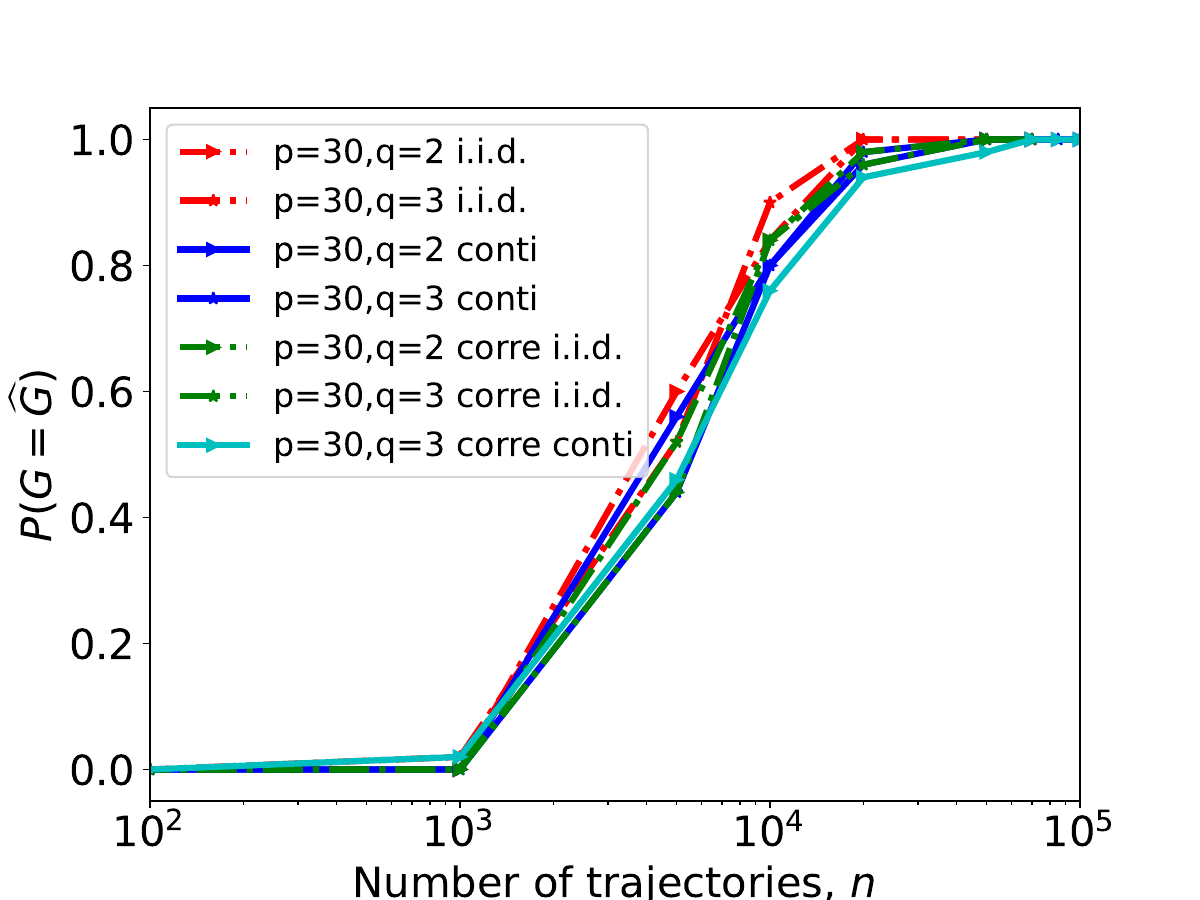}
 \caption{$p=30$}
 \end{subfigure}
 \hfill
\caption{Probability of error with number of trajectories, for different networks under restart and record (RR) and continuous (conti) sampling strategies, when the system is excited by either i.i.d. or WSS noise. $p$ and $q$ refer to the number of nodes and degree of each node in the network, respectively. The trajectory length $N=64$ is fixed for each trajectory, and $\omega=\frac{17}{64}$.}
\label{fig:plots}
\end{figure*}

The following Lemma shows that $Q_{approx}$ is small if $N$ (length of each trajectory) is large.
\begin{lem}[Lemma 5.1, \cite{doddi2022efficient}]
\label{lem:Spectrogram_error_finite_data}
Consider an LDS given by \eqref{eq:LDS} that satisfies Assumption \ref{ass:autocorrelation}. Let $Q_{approx}=\widetilde{\Phi}_x(\omega)-{\Phi}_x(\omega)$ where $\widetilde{\Phi}_x(\omega)$ is given in Eq.~\ref{eq:DFT_covariance}. Then $\|Q_{approx}\|<\varepsilon_1$ if $N > \frac{2C\rho^{-1}}{(1-\rho^{-1})^2\varepsilon_1}$. 
\end{lem}
The next step is to characterize $Q_2$, the error in estimating $\widetilde{\Phi}_\x$ using $\hatt{\Phi}_x(\omega)$. Since $\mathbb{E}\left(\x^r(\omega)[\x^r(\omega)]^*\right)=\widetilde{\Phi}_x(\omega))$, $\hatt{\Phi}_x(\omega)$ is an unbiased estimator of $\widetilde{\Phi}_x(\omega)$. The following theorem provides a concentration bound on $\|Q_2\|$. The concentration bound is applicable under two sampling scenarios; the restart and record setting and continuous sampling setting, as shown in Fig.~\ref{fig:plots}. 
\begin{thm}
\label{thm:Q_concentration_bound}
Suppose $\{\breve{x}^r(k)\}_{k=0}^{N-1}$, $1\leq r\leq n$ be the time series measurements obtained from an LDS governed by \eqref{eq:LDS}, satisfying Assumption \ref{ass:psdm}. Then $\forall \omega \in \Omega$,
\begin{align}
\label{eq:Q_concentration_bound}
 \mP\left( \left\|\hatt{\Phi}_x(\omega)-\widetilde{\Phi}_x(\omega))\right\|\geq \epsilon \right) \leq \exp \left(-\frac{\epsilon^2n}{128 M^2}+6p \right). 
\end{align}

\end{thm}
The detailed proof is provided in the Supplementary material, Section 3. We provide a concise proof sketch below.

\textbf{\underline{Proof sketch of Theorem \ref{thm:Q_concentration_bound}}:} The first step is to upper bound the spectral norm using a finite $\delta$-cover of unit sphere $\mbS^p$, that is, $\|\Q\|:=\sup_{v \in \mbS^p} v^*\Q v \leq 2 \max_{j=1,\dots,m} |(v^j)^*Qv^j| $, where $v^j$ denotes the center of the $j$th sphere in the finite spherical cover. We then upper bound the moment-generating function, $\mbE\left[ e^{\lambda \|Q\|} \right]\leq \sum _{j=1}^m\mbE\left[ e^{2\lambda(v^j)^*Qv^j} \right]+\mbE\left[ e^{-2\lambda(v^j)^*Qv^j} \right]$. The upper bound on $\mbE\left[ e^{-2\lambda(v^j)^*Qv^j} \right]$ is divided into two parts. We first show the proof for the restart and record sampling by exploiting the properties of i.i.d. nature of the Fourier transform of the samples. Next, we generalize the results to the continuous sampling settings. The proofs involve using the Rademacher random variable based symmetrization argument followed by a series of standard inequalities. Finally, Chernoff bounding technique is applied to the upper bound of the moment-generating function.
\qed

By combining Lemma \ref{lem:Spectrogram_error_finite_data} and Theorem \ref{thm:Q_concentration_bound}, the following corollary is obtained, which gives a concentration bound on the estimation error, $\|Q\|$.
\begin{cor}
\label{cor:PSD_estimation_error}
  Consider an LDS governed by \eqref{eq:LDS} that satisfies Assumptions \ref{ass:psdm} and \ref{ass:autocorrelation}. Let $\{\breve{x}^r(k)\}_{k=0}^{N-1}$, $1\leq r\leq n$ be the time series measurements obtained for the LDS. Suppose that $N > \frac{2C\rho^{-1}}{(1-\rho^{-1})^2\varepsilon_1}$, where $0<\varepsilon_1$. Let $0<\varepsilon_1,\varepsilon_2<\varepsilon$ be such that $\varepsilon_2=\varepsilon-\varepsilon_1$. Then $\forall \omega \in \Omega$,
 \begin{align}
\label{eq:cor:Q_bound}
 \mP\left( \left\|{\Phi}_{x}(\omega)-\hatt{\Phi}_{x}(\omega))\right\|\geq \varepsilon \right) &\leq \exp \left(-\frac{\varepsilon_2^2n}{128 M^2}+6p \right). 
\end{align}
\end{cor}

\subsection{Sample Complexity Bounds: Upper Bound}
In the previous subsection, concentration bounds on the estimation errors in PSDM were obtained. Here, a concentration bound on the error in estimating $f$ is obtained, which is used to obtain a concentration bound in reconstructing the DDAG, $G$. The following result provides a concentration bound on $|f-\hatt{f}|$.
\begin{lem}
\label{lem:f_concentration_bound}
 Consider an LDS governed by \eqref{eq:LDS} that satisfies Assumptions \ref{ass:psdm} and \ref{ass:autocorrelation}. Let $\{\breve{x}^r(k)\}_{k=0}^{N-1}$, $1\leq r\leq n$ be the time series measurements obtained for the LDS. Suppose that $N > \frac{2C\rho^{-1}}{(1-\rho^{-1})^2\varepsilon_1}$, where $0<\varepsilon_1$. Let $0<\varepsilon_1,\varepsilon_2<\varepsilon$ be such that $\varepsilon_2=\varepsilon-\varepsilon_1$. Then there exists a $c_0 \in \mbR$ such that, for any $\omega \in \Omega$,
 \begin{align*}
 \mP\left( | f(i,C,\omega)-\hatt{f}(i,C,\omega)\geq \varepsilon \right) &\leq c_0e^{\left( -\frac{\varepsilon_2^2n}{10368 M^{6}}+6(q+1)\right)},
\end{align*}
where $q$ is the maximum number of parents any node has in $G$.
\end{lem}
The proof, provided in Section 4 of the Supplementary material, uses Corollary \ref{cor:PSD_estimation_error} for different entries in the formula for $f(i,C,\omega)$ as defined in \eqref{eq:f}.
Based on Lemma \ref{lem:f_concentration_bound}, the following upper bound on the probability of error in estimating $G$ can be obtained. \begin{thm}
\label{thm:G_upperbound}
Suppose $q \leq p/2$. Consider an LDS that belongs to $\cH_{p,q}(\beta,\sigma,M)$ (Definition \ref{def:H_p,q}) that satisfies Assumptions \ref{ass:psdm} and \ref{ass:autocorrelation}. Let $\{\breve{x}^r(k)\}_{k=0}^{N-1}$, $1\leq r\leq n$ be the time series measurements of the LDS and let $\hatt{G}$ be the DDAG reconstructed by Algorithm \ref{alg:Ordering}. Suppose that $N > \frac{2C\rho^{-1}}{(1-\rho^{-1})^2\varepsilon_1}$, where $0<\varepsilon_1<\Delta/4$. Let $0<\varepsilon_1,\varepsilon_2<\varepsilon<\Delta/4$ be such that $\varepsilon_2=\varepsilon-\varepsilon_1$.
Then $\mP\left(G(\omega) \neq \hatt{G}(\omega)\right)\leq \delta$ if $$n \gtrsim \frac{M^6\left(q\log(p/q)-\log \delta\right)}{\epsilon_2^2}. $$
\end{thm}\begin{proof}
See supplementary material, Section 5.
\end{proof}

In the following section, a matching lower bound is derived.

\section{Sample Complexity Bounds: Lower Bound}
The lower bound for reconstructing DDAG is derived using information-theoretic techniques, in particular Fano's inequality, and by restricting the interested family of graphs to a finite set. Notice that, except for a couple of non-trivial facts in dynamic setup, this is a direct extension of the lower bound for the static case provided in \cite{gao2022optimal}. The approach is to construct restricted ensembles of graphical models and then to lower bound the probability of error using Generalized Fano's inequality. Theorem \ref{thm:Lower_bound} below provides the lower bound.
For completeness, the proof is provided in the supplementary material.

\begin{thm}
\label{thm:Lower_bound}
Suppose $q \leq p/2$. If 
\begin{align*}
n \leq (1-2\delta) \max\left( \frac{ \log p}{2\beta^2+ \beta^4},\frac{ q\log (p/q)}{M^2-1} \right)
\end{align*}
 then 
 \begin{align*}
 \inf_{\widehat{G}} \max_{H \in \cH_{p,q}(\beta,\sigma,M)} \mP\{ (G(H)\neq \hatt{G})\} \geq \delta,
 \end{align*}
\end{thm}
That is, if $n \leq (1-2\delta) \max\left( \frac{ \log p}{2\beta^2+ \beta^4},\frac{ q\log (p/q)}{M^2-1} \right)$, then any given estimator fails to reconstruct the DDAG with probability greater than $\delta$. The lower-bound provides the fundamental limit on reconstructing the DDAG from finite samples.

Theorem \ref{thm:Lower_bound} provides the lower bound $\Omega\left(\frac{ \log p}{2\beta^2+ \beta^4} \bigvee \frac{ q\log (p/q)}{M^2-1}\right)$. Notice that the upper bound in Theorem \ref{thm:G_upperbound} is $O(q\log\frac{p}{q})$. Thus we obtain a matching order bound when the lower bound is dominated by the second term.

\section{Simulation Experiments}
The effectiveness of Algorithm \ref{alg:Ordering} is demonstrated with the help of a simulation study here. In the simulations, we fix the number of nodes, $p$, and the number of parents, $q$ as in \cite{gao2022optimal}. The exogenous noise PSD is fixed to be the same for all the nodes. To generate the random DAGs, first obtain a random permutation of $[p]$ nodes to obtain an ordering $\tau$. Then for each $j \in [p]$, randomly select $\min(q,j-1)$ number of parents from $\tau_{[1:j-1]}$. Then for each edge $(j,i)$, generate $B_{ij} \sim \epsilon \times Unif(.5,1)$, where $\epsilon$ is the Rademacher random variable. The matrix $B$ is then scaled by $\frac{1}{1.002\|B\|}$ to satisfy Assumption \ref{ass:autocorrelation}. The data is generated according to the model 
\begin{align}
\label{eq:AR_generator}
 x_i(t)=\sum_{j\neq i}B_{ij} x_j(t-1)+e_i(t), \forall i \in [p],t\in [T]
\end{align}
We consider two scenarios for $e_i(t)$, viz 1) zero mean i.i.d. Gaussian noise with variance $\sigma=.5$, and 2) temporally correlated WSS Gaussian noise. The correlated WSS noise is generated according to the model $\e(t)=\alpha\e(t-1)+\w(t)$, for every $t \in [T]$, with $\alpha=0.5$ and $\w(t)$ i.i.d. Gaussian noise with covariance matrix $0.5 \I$. The total number of samples, $T=N \times n$, where $N$ is the length of each trajectory and $n$ is the number of trajectories. We perform both restart and record and continuous sampling, as detailed in Fig.~\ref{fig:plots}. 

For each $p$ and $q$, we generated 50 random networks and time series data according to the model above. For each set of parameters, $n$, $T$ and $N$, we applied Algorithm \ref{alg:Ordering} along with the estimator from Section \ref{sec:DFT_PSD_est} for $\omega=\frac{17}{64}$ to reconstruct the graph. Then, $\mP(G=\hatt{G})$ is empirically given by by $\#(G=\hatt{G})/50$. Fig. \ref{fig:plots} show the comparison for various test parameters. The simulations are performed using Python in a Macbook with M1 pro chip and 16 GB RAM. The codes are available {\href{https://github.com/Mishfad/Learning-Dynamical-DAGs}{here}}.

\section*{Conclusion}
In this article, we characterized the optimal sample complexity for structure identification with directions in linear dynamical networks. Inspired by the static setting, a metric, and an algorithm were proposed based on the power spectral density matrix to exactly reconstruct the DAG. It is shown that the optimal sample complexity is $n=\Theta(q\log (p/q))$. For the upper bound characterization, we obtained a tight concentration bound for the power spectral density matrix. An information-theoretic min-max lower bound also was provided for (sub) Gaussian linear dynamical systems. It was shown that the upper bound is order optimal with respect to the lower bound.

\section*{Acknowledgement}
The authors acknowledge support from the Information Science and Technology Institute (ISTI) and the Directed Research and Development project: “Resilient operation of interdependent
engineered networks and natural systems” at Los Alamos National Laboratory. The first and the third authors acknowledge the support of NSF through Award Number 2030096 and ARPA-E through the Award number DE-AR0001016.The research work conducted at Los Alamos National Laboratory is done under the auspices of the National Nuclear Security Administration of the U.S. Department of Energy under Contract No. 89233218CNA000001. 
\bibliography{ref}
\bibliographystyle{abbrvnat}

\newcommand{\aistatssupptitle}[1]{
 \toptitlebar 
 \begin{center}
 {\Large\bfseries #1 \par} 
 \end{center}
 \bottomtitlebar 
}

\onecolumn
\aistatssupptitle{Supplementary Material}
\setcounter{section}{0}
\section{Proof of Lemma 3.3}
\label{app:lem:f_comparison}
Let $C \subseteq V \setminus \{j\}$ be an ancestral set and let $D=nd(j) \setminus C$. Then, $$\x_j(\omega)=H_{jC}(\omega)X_C(\omega)+H_{jD}(\omega)X_D(\omega)+e_j(\omega).$$
Applying $\Phi_{e_j C}=\Phi_{e_j D}=0$, we obtain $\Phi_{jC}(\omega)=H_{jC}(\omega) \Phi_{CC}(\omega)+H_{jD}(\omega) \Phi_{DC}(\omega)$ and
{\footnotesize\begin{align*}
 \Phi_{j}(\omega)&=H_{jC}(\omega)\Phi_CH_{jC}(\omega)^*+H_{jD}(\omega) \Phi_{DC}(\omega) H_{jC}^*(\omega)+H_{jC}(\omega) \Phi_{CD}(\omega)H_{jD}^*(\omega)+H_{jD}(\omega) \Phi_{D}(\omega)H_{jD}^*(\omega)+\Phi_{e_je_j}(\omega).
\end{align*} }
Then 
\begin{align*}
f(j,C,\omega)=\Phi_{j}-\Phi_{jC}\Phi_{C}^{-1}\Phi_{Cj}
&=\Phi_{e_j e_j}+H_{jD} (\Phi_{D}- \Phi_{DC}\Phi_{CC}^{-1}\Phi_{CD})H_{jD}^*.
\end{align*}
Notice that when $Pa(j) \subseteq C$, $H_{jD}=0$, and $f(j,C,\omega)=\Phi_{e_j e_j}$, which shows the first part.

To prove the second part, suppose $Pa(j) \cap D \neq \emptyset$. We need to show that $H_{jD} (\Phi_{D}- \Phi_{DC} \Phi_{CC}^{-1} \Phi_{CD}) \H_{jD}^* >0$. Let $A=nd(j) = C\cup D$ and $B=desc(j)\cup \{j\}$. From \cite{Talukdar_ACC18,doddi2019topology},
\begin{align*}	
\Phi_{AA}^{-1}=&\S+\L,\text{ where }\\
\S&=(\I_A-\H_{AA}^*)\Phi_{e_A}^{-1}(\I_A-\H_{AA}),\\
\L &=\H_{BA}
\Phi_{e_B}^{-1}\H_{BA}-\Psi^*\Lambda^{-1}\Psi,\\ \nonumber
\Psi&=\H_{AB}^*\Phi_{e_A}^{-1}(\I-\H_{AA})+(\I-\H^*_{BB})\Phi_{e_B}^{-1}\H_{BA}, \text{ and}\\ \nonumber
\Lambda&=\H_{AB}^*\Phi_{e_A}^{-1}\H_{AB}+(\I-\H^*_{BB})\Phi_{e_B}^{-1}(\I-\H_{BB}).
\end{align*}
Notice that since $B$ is the set of descendants of $j$, $\H_{AB}=0$, as cycles can be formed otherwise. Then, $\L=0$ and $\Phi_{AA}^{-1}=(\I_A-\H_{AA}^*)\Phi_{e_A}^{-1}(\I_A-\H_{AA})$. 

\begin{align*}
\Phi_{AA}^{-1}=\left[ \begin{array}{cc}
 \Phi_{DD} & \Phi_{DC} \\
 \Phi_{CD} & \Phi_{CC} 
\end{array} \right]^{-1} =\left[ \begin{array}{cc}
 K_{DD} & K_{DC} \\
 K_{CD} & K_{CC} 
\end{array} \right]=\frac{1}{\sigma}\left(\I-\H_{AA}^*\right)\left(\I-\H_{AA}\right) 
\end{align*}
By Scur's complement, $(\Phi_{D}- \Phi_{DC}\Phi_{CC}^{-1}\Phi_{CD})^{-1}=K_{DD}=\frac{1}{\sigma}(\I_D-\H^*_{DD}-\H_{DD}+(\H_{AA}^*\H_{AA})_{D\times D})$. Moreover, $$\H_{AA}=\left[ \begin{array}{cc}
 \H_{DD} & \H_{DC} \\
 \H_{CD} & \H_{CC} 
\end{array} \right] \text{ and } (\H_{AA}^*\H_{AA})_{D\times D}=\H_{DD}^*\H_{DD}+ \H_{CD}^*\H_{CD}.$$ Since $C$ is ancestral, $\H_{CD}=0$ and $$K_{DD}=\frac{1}{\sigma}(\I_D-\H_{DD})^*(\I_D-\H_{DD}).$$

Since $G$ is a DAG, the rows and columns of $\H$ can be rearranged to obtain a lower triangular matrix with zeros on the diagonal. Thus eigenvalues of $(\I_D-\H_{DD})$ and its inverse are all ones. Hence minimum eigenvalue of $K^{-1}_{DD}$ is greater than $\sigma$.
Applying Rayleigh Ritz theorem on $\H_{jD}K^{-1}_{DD}\H_{jD}^*$, we have 
\begin{align}
\label{eq:KDD_proof_sum} \H_{jD} (\Phi_{D}- \Phi_{DC}\Phi_{CC}^{-1}\Phi_{CD})\H_{jD}^* =\H_{jD}K^{-1}_{DD}\H_{jD}^*\geq \sigma|D| \beta^2
\end{align}
which is strictly greater than zero if $D$ is non-empty. \hfill 
\qed

\section{Proof of Lemma 3.8}
\label{app:lem:G_reconstruction}
 The proof is done in two steps. First, we show that $\cS$ in Algorithm 1 is a topological ordering. Then, we show that step (4) in Algorithm 1 can identify the parents of every node in $G$. The first step is shown via induction. Since $|\hatt{f}(i,C,\omega)-{f}(i,C,\omega)| < \Delta/4$ for empty set, $|\Phi_{ii}-\hatt{\Phi}_{ii}|<\Delta/4$ for every $i$. Recall from Lemma 3.3 that $\Phi_{ii}-\Phi_{jj}> \Delta$ if $ i$ is a source node and $j$ is a non-source node. Then,
$\hatt{\Phi}_{jj} \geq {\Phi}_{jj}-\Delta/4\geq \Phi_{ii}+3\Delta/4 \geq \hatt{\Phi}_{ii}+\Delta/2$. Thus, $i \in \arg \min \limits_{1 \leq k \leq p} \hatt{\Phi}_{kk}$ if and only if $i \in \arg \min \limits_{1 \leq k \leq p} {\Phi}_{kk}$ and thus $\cS_1$ is always a source node.

 For the induction step, assume that $\cS_1,\dots,\cS_n$ forms a correct topologically ordered set w.r.t. $G$. Let $C \subseteq \cS(1:n)$. If $Pa(i) \subseteq C$ and $Pa(j) \nsubseteq C$, then by applying Lemma 3.3,
$\hatt{f}(j,C,\omega) > {f}(j,C,\omega)-\Delta/4 \geq \sigma+3\Delta/4=f(i,C,\omega)+3\Delta/4$ $\geq \hatt{f}(i,C,\omega)+\Delta/2$. Thus, $i \in \arg \min \limits_{k \in V \setminus \cS} \hatt{f}(k,C,\omega)$ if and only if $i \in \arg \min \limits_{k \in V \setminus \cS}{f}(k,C,\omega)$ and thus $(\cS,\cS_{n+1})$ forms a topological order w.r.t. $G$, by Lemma 3.6.

To prove the second step, let $C \subseteq \cS(1:i)$. Since $\cS(1:i)$ is a valid topological ordering, $Pa(i) \subseteq \cS(1:i-1)$. Let $k \in Pa(i)$ and let $D=C\setminus \{k\}$. Then, as shown in Corollary 3.4 $f(i,C,\omega)-f(i,D,\omega)\geq \Delta$, and
\begin{align*}
 \Delta \leq |f(i,C,\omega)-f(i,D,\omega)|&\leq |f(i,C,\omega) - \hatt{f}(i,C,\omega)|+|\hatt{f}(i,C,\omega)-\hatt{f}(i,D,\omega)|\\
 & \hspace{1cm}+|\hatt{f}(i,D,\omega)-f(i,D,\omega)|\\
 &< \Delta/4+ \Delta/4+|\hatt{f}(i,C,\omega)-\hatt{f}(i,D,\omega)|\\
 \implies |\hatt{f}(i,C,\omega)-\hatt{f}(i,D,\omega)| &> \Delta/2.
\end{align*}
Suppose $k \notin Pa(i)$ but $k \in \cS(1:i)$. Then, for $D=C\setminus \{k\}$, $f(i,C,\omega)-f(i,D,\omega)=0$. Repeating the same series of inequalities above by exchanging $f$ and $\hatt{f}$, we obtain $|\hatt{f}(i,C,\omega)-\hatt{f}(i,D,\omega)|<\Delta/2$.

Thus, from the set $\cS$, for every node $\cS_i$, one can check nodes $\cS_1,\dots,\cS_{i-1}$ and verify if the difference of including and excluding the node is greater than $\Delta/2$. If the difference is greater than $\Delta/2$ for some $k$, then $k$ is a parent of $i$, and if not, then the node is not a parent of $i$. That is,
let $C_i=\{\cS_1,\dots,\cS_{i-1}\}$, $i>1$, and let $$\hatt{P}_i:=\left\{ j \in C_i \left| ~ |\hatt{f}(\cS_i,C_i,\omega) - \hatt{f}(\cS_i,C_i \setminus \{j\},\omega) | > \Delta/2 \right. \right\}.$$ Then, $Pa(i)=\hatt{P}_i$.
\hfill\qed
\section{Proof of Theorem 4.2}
\label{app:thm:Q_concentration_bound}

By the variational form of spectral norm \cite{matrix_analysis}, $$\|Q\|=\sup_{v \in \mathbb{C}^{p}, \|v\|=1}|v^*Qv|,$$ where the max is taken over a $p$-dimensional unit complex sphere, $\mathbb{S}^{p}:=\{v \in \mathbb{C}^{p}: \|v\|_2=1\}$. The first step here is to reduce supremum to finite maximization using finite covers of a unit ball, which is done using a $\delta$ cover. A $\delta$-cover of a set $\mathcal{A}$ is a set $v^1,\dots,v^m$ such that for every $v \in \mathcal{A}$, there exists an $i \in 1,\dots,m$ such that $\|v^i-v\|_2 \leq \delta$. The following Lemma is obtained by extending example 5.8 in \cite{wainwright_2019} to the complex field.

\begin{lem}
\label{lem:delta_cover}
Let $v^1,\dots,v^m$ be a $\delta$-covering of the unit sphere $\mathbb{S}^{p}$. Then there exists such a covering with $m\leq (1+2/\delta)^{2p}$ vectors.
\end{lem}
\begin{proof}
 The proof follows by extending (5.9) in \cite{wainwright_2019}, to the complex field.
\end{proof}
Let $v\in \mbS^{p}$ and let $v^j$ be such that $v=v^j+\Delta$, where $\|\Delta\| \leq \delta$. Then, $v^*Qv=(v^j)^*Qv^j+2 \Re\{\Delta^*Qv^j\}+\Delta^*Q\Delta 
$. Applying triangle inequality,
\begin{align*}
 |v^*Qv| &\leq |(v^j)^*Qv^j|+2\|\Delta\|\|Q\|\|v^j\|+|\Delta\|^2\|Q\|\\
 &\leq |(v^j)^*Qv^j|+2 \delta\|Q\|+\delta^2\|Q\|\\
 &\leq |(v^j)^*Qv^j|+\frac{1}{2}\|Q\| \text{ for } \delta\leq 0.22474.
\end{align*}
Thus,
\begin{align*}
\|Q\|&=\max_{v \in \mbS^{p}}|v^*Qv| \leq \max_{j=1,\dots,m} |(v^j)^*Qv^j|+\frac{1}{2}\|Q\| \text{ and }\\
\|Q\| &\leq 2 \max_{j=1,\dots,m} |(v^j)^*Qv^j|
\end{align*}
Next, we find an upper bound for $\mbE\left[ e^{\lambda \|Q\|} \right]$, which is treated with Chernoff-type bounding technique to obtain the desired result.
\begin{align}
\nonumber 
\mbE\left[ e^{\lambda \|Q\|} \right] &\leq \mbE\left[ \exp\left({2 \lambda \max_{j=1,\dots,m} |(v^j)^*Qv^j|}\right) \right]\\
\label{eq:exp_lambda_Q_sum}
&\leq \sum _{j=1}^m\mbE\left[ e^{2\lambda(v^j)^*Qv^j} \right]+\mbE\left[ e^{-2\lambda(v^j)^*Qv^j} \right]
\end{align}
Next, we complete the proof for the restart and record sampling and the continuous sampling separately.
\subsection{Restart and Record Sampling}
Under the restart and record sampling settings, for any given $\omega\in \Omega$, $\{\x^r(\omega)\}_{r=1}^n$ is i.i.d. Thus
\begin{align*}
 \mbE\left[ \exp\left({t(v^j)^*Qv^j}\right) \right]&=\mbE\left[ \exp\left({t(v^j)^*(\hatt{\Phi}_x(\omega)-\widetilde{\Phi}_x(\omega)))v^j}\right) \right]\\
 &=\mbE\left[ \exp\left({\frac{t}{n} \sum_{r=1}^n(v^j)^*\x^r(\omega)[\x^r(\omega)]^*v^j-(v^j)^*\widetilde{\Phi}_x(\omega)v^j}\right) \right]\\
 &=\prod_{r=1}^n\mbE\left[ \exp\left({\frac{t}{n} (v^j)^*\x^r(\omega)[\x^r(\omega)]^*v^j-(v^j)^*\widetilde{\Phi}_x(\omega)v^j}\right) \right]\\
 &=\left(\mbE\left[ \exp\left({\frac{t}{n} (v^j)^*\x^1(\omega)[\x^1(\omega)]^*v^j-(v^j)^*\widetilde{\Phi}_x(\omega)v^j}\right) \right]\right)^n\\
 &=\left(\mbE\left[ \exp\left({\frac{t}{n} |v^*\x^r(\omega)|^2-v^*\widetilde{\Phi}_x(\omega)v}\right) \right]\right)^n
\end{align*}
 Let $\varepsilon \in \{-1,+1\}$ be a Rademacher variable independent of $\x^r$. It can be shown that Proposition 4.11 in \cite{wainwright_2019} will hold for complex numbers also. Then 
\begin{align}
\label{recall1}
 \mbE_{\x^r(\omega)}\left[ \exp\left({\frac{t}{n} |v^*\x^r(\omega)|^2-v^*\widetilde{\Phi}_x(\omega)v}\right) \right] &\leq \mbE_{\x^r(\omega),\varepsilon}\left[ \exp\left({\frac{2t \varepsilon}{n} |v^*\x^r(\omega)|^2}\right) \right]\\
 &= \sum_{k=0}^\infty \frac{\left(2t/n\right)^{2k}}{2k!} \mathbb{E} \left[ |v^*\x^r(\omega)|^{4k}\right]
\end{align}

Recall that $\widetilde{\Phi}_\x $ is a positive definite matrix and $v^*\x^r \sim N(\mathbf{0}, \eta)$, where $\eta=v^*\widetilde{\Phi}_\x v \leq \lambda_{max}(\widetilde{\Phi}_\x)\leq M.$ The even moments of $y\sim N(0,\eta)$ is given by $\mbE\{y^{2k}\}=\eta^{2k} (2k-1)!!= \frac{(2k)!}{ 2^{k}k!}\eta^{2k}
$. 
Then
$$\mathbb{E} \left[ |v^*\x^r(\omega)|^{4k}\right] \leq \frac{(4k)!}{2^{2k}(2k)!} M^{2k}.$$
Therefore using the inequality $(4k)! \leq 2^{2k}[(2k)!]^2$,
\begin{align*}
 \mbE_{\x^r(\omega)}\left[ \exp\left({\frac{t}{n} |v^*\x^r(\omega)|^2-v^*\widetilde{\Phi}_x(\omega)v}\right) \right] &\leq 1+\sum_{k=1}^\infty \frac{\left(2t/n\right)^{2k}}{2k!} \frac{(4k)!}{2^{2k}(2k)!} M^{2k}\\
 &\leq 1+\sum_{k=1}^\infty \frac{\left(2t/n\right)^{2k}}{2k!} \frac{2^{2k}[(2k)!]^2}{2^{2k}(2k)!} M^{2k}\\
 &= 1+\sum_{k=1}^\infty \left(\frac{2Mt}{n}\right)^{2k} = \frac{1}{1-\left(\frac{2Mt}{n}\right)^2}\\ 
 & \leq \exp \left( \frac{8M^2t^2}{n^2}\right)
\end{align*}
whenever $\frac{2Mt}{n} < 3/4$, where the final inequality follows by applying $1-x \geq e^{-2x}$ for $x \in [0,3/4]$ (to be precise 0.77). 
Thus, $$\mbE\left[ \exp\left({t(v^j)^*Qv^j}\right) \right] \leq \exp \left( \frac{8M^2t^2}{n}\right),~ \forall |t| \leq \frac{3n}{8M}.$$
Applying Lemma \ref{lem:delta_cover} and the bound $2m \leq 2 (1+2/0.22474)^{2p} \leq 2e^{4.6p}\leq e^{5p+0.693}\leq e^{6p},$

\begin{align*}
\text{From \eqref{eq:exp_lambda_Q_sum},}\quad \mbE\left[ e^{\lambda \|Q\|} \right] &\leq \mbE\left[ \exp\left({2 \lambda \max_{j=1,\dots,m} |(v^j)^*Qv^j|}\right) \right]\\
 &\leq \sum _{j=1}^m\mbE\left[ e^{2\lambda(v^j)^*Qv^j} \right]+\mbE\left[ e^{-2\lambda(v^j)^*Qv^j} \right]\\
 &\leq 2m \exp \left( \frac{32M^2\lambda^2}{n}\right)\\
 &\leq \exp \left( \frac{32M^2\lambda^2}{n} + 6p\right), ~\forall |\lambda| \leq \frac{3n}{16M}.
\end{align*}
 Applying Chernoff-type bounding approach,
\begin{align*}
 \mP\left( \|Q\|\geq t \right) \leq e^{-\lambda t} \mbE\left[ e^{\lambda \|Q\|} \right] \leq \exp \left( -\lambda t +\frac{32M^2\lambda^2}{n} + 6p \right), ~ \forall |\lambda| \leq \frac{3n}{16M}. 
\end{align*}
The tightest bound is given by $g^*(t):= \inf \limits_{|\lambda| \leq \frac{3n}{16M}} \left\{-\lambda t +\frac{32M^2\lambda^2}{n} + 6p\right\}$, where the objective is convex. Taking derivative w.r.t. $\lambda$ and equating to zero, $\lambda^*=\frac{tn}{64 M^2}$ and $g^*=-\frac{t^2n}{64 M^2}+ \frac{32 M^2}{n} \frac{t^2n^2}{64^2 M^4}+6p$ $=6p-\frac{t^2n}{128 M^2}$, if $t$ is such that $t \leq 12M$, which is reasonable as we can always pick $M\geq 1$. 

Thus, $\mP\left( \left\|\Q\right\|\geq t \right) \leq \exp \left(-\frac{t^2n}{128 M^2}+6p \right)$
The theorem statement follows. \hfill\qed
\subsection{Continuous Sampling} 
In the continuous sampling setting, the samples $\breve{\x}(0),\dots,\breve{\x}(N-1),\breve{\x}(N),\dots,\breve{\x}(2N-1),\dots,\breve{\x}((n-1)N),\\\dots,\breve{\x}(nN-1)$ are sampled continuously and are correlated with each other. Thus, $\x^r(\omega)$ and $\x^s(\omega)$, $r\neq s,~1\leq r,s\leq n$, can be correlated, in contrast to the restart and record (RR) setting, where the 
$\x^r(\omega)$ and $\x^s(\omega)$, $r\neq s$ are i.i.d. For any given $\omega \in \Omega$, let $\x(\omega):=[[\x^{1}(\omega)]^T,[\x^{2}(\omega)]^T,\dots,[\x^n(\omega)]^T]^T\in \mC^{pn \times 1}$ be the vectorized form of $\{\x^r(\omega)\}_{r=1}^n$ and let $\cC(\omega):= \mbE\{\x(\omega)\x^*(\omega)\}$ be the covariance matrix of $\x(\omega)$. 
Under the RR setting, $\cC(\omega) \in \mC^{pn \times pn}$ will be a block-diagonal matrix (of block size $p \times p$), whereas in the continuous sampling, the non-block-diagonal entries of $\cC(\omega)$ can be non-zero. However, the vector $\x(\omega)$, with correlated entries, can be written as a linear transformation of i.i.d. vector $\w \in \mC^{pn \times 1}$ with unit variance, i.e. $\x(\omega)=\cC^{1/2}(\omega)\w$, where $\cC^{1/2}$ is the square-root of $\cC$ and $\w \sim \mathcal{N}(\mathbf{0},\I_{pn})$. It follows from \cite{doddi2022efficient} that $\x^r(\omega)$ and thus $\x(\omega)$ are Gaussian distributed when $\{\bre(k)\}_{k=1}^n$ in the linear time-invariant model (1) are Gaussian. It can be verified that $\mbE\{\x(\omega)\x^*(\omega)\}=\cC^{1/2}(\omega)\mbE\{\w\w^*\}\cC^{1/2}(\omega)=\cC(\omega)$. Notice that the covariance matrix $\cC(\omega)$ is a block matrix, defined as

\begin{align*}
 \cC=\begin{bmatrix}
 \cC^{11} & \cC^{12}&\dots&\cC^{1n}\\
 \cC^{21} & \cC^{22}&\dots&\cC^{2n}\\
 \vdots\\
 \cC^{n1} & \cC^{n2}&\dots&\cC^{nn}
 \end{bmatrix}, \text{where $\cC^{rs}(\omega) \in \mC^{p\times p}$, } 1\leq r,s\leq n,
\end{align*}
where the entries of $\cC^{rs}(\omega)$ is given by $\mathbb{E}\left\{\x^r(\omega)[\x^s(\omega)]^*\right\}$. Recall that $$\x^r(\omega)=\frac{1}{\sqrt{N}} \sum_{\ell=0}^{N-1} \breve{x}((r-1)N+\ell) e^{-i\omega \ell}.$$ 

Let $\I_r=\begin{bmatrix}
 \mathbf{0}|\dots |\mathbf{0}|\I_{p \times p}|\dots |\mathbf{0}
\end{bmatrix} \in \mbR^{p \times np}$ be such that $r^{th}$ block is identity matrix. Then $\x^r(\omega)=\I_r \x(\omega)$. The estimated PSDM is then given by
\begin{align*}
 \hatt{\Phi}_\x(\omega)=\frac{1}{n} \sum_{r=1}^n \x^r(\omega) [\x^r(\omega)]^*=\frac{1}{n}\sum_{r=1}^n \I_r\x(\omega)\x^*(\omega)\I_r^*. 
\end{align*}~Substituting $\x(\omega)=\cC^{1/2}(\omega)\w$, and letting $\B(\omega):=(\cC^{1/2})^*(\omega)\sum_{r=1}^n\I_r^* u u^*\I_r\cC^{1/2}(\omega)$ 

\footnotesize
\begin{align*}
 \mbE\left[ \exp\left({tu^*\Q u}\right) \right] &{=}\mbE\left[ \exp\left({\frac{t}{n} \sum_{r=1}^n u^*\x^r(\omega) [\x^r(\omega)]^*u}-u^*\widetilde{\Phi}_x(\omega)u\right) \right]\\
 &{=} \mbE\left[ \exp\left({\frac{t}{n}\sum_{r=1}^n\left[ \x^*(\omega)\I_r^* u u^*\I_r\x(\omega) - \mbE\left\{ \x^*(\omega)\I_r^* u u^*\I_r\x(\omega) \right\} \right] }\right) \right]\\
 &{=} \mbE\left[ \exp\left({\frac{t}{n}\sum_{r=1}^n\left[ \w^*(\cC^{1/2})^*(\omega)\I_r^* u u^*\I_r\cC^{1/2}(\omega)\w - \mbE\left\{ \w^*(\cC^{1/2})^*(\omega)\I_r^* u u^*\I_r\cC^{1/2}(\omega)\w \right\} \right] }\right) \right]\\
&{=} \mbE\left[ \exp\left({\frac{t}{n}\left[ \w^*(\cC^{1/2})^*(\omega)\sum_{r=1}^n(\I_r^* u u^*\I_r)\cC^{1/2}(\omega)\w - \mbE\left\{ \w^*(\cC^{1/2})^*(\omega)\sum_{r=1}^n(\I_r^* u u^*\I_r)\cC^{1/2}(\omega)\w \right\} \right] }\right) \right]\\
&=\mbE\left[ \exp\left({\frac{t}{n}\left[ \w^*\B(\omega)\w - \mbE\left\{ \w^*\B(\omega)\w \right\} \right] }\right) \right]\\
\end{align*}
\normalsize
{Notice that $\I_r^*u=\begin{bmatrix} {0}, \dots, {0}, u^T,,\dots,{0} \end{bmatrix}^T$ is a column vector, 
{\footnotesize$$ \I_r^* u u^*\I_r=\begin{bmatrix}
 \mathbf{0} &\mathbf{0} & \mathbf{0}& \dots&\mathbf{0}\\
 & \dots & &\dots&\mathbf{0}\\
 \vdots & \dots&\underbrace{uu^*}_{(r,r)^{th} ~block} & \dots& \mathbf{0}\\
 \vdots & \vdots& \cdots&\ddots&\mathbf{0}\\
 \mathbf{0}& \mathbf{0}&\dots& \mathbf{0}
\end{bmatrix} \text{ and } \sum_{r=1}^n\I_r^* u u^*\I_r=\begin{bmatrix}
 uu^* & \mathbf{0}& \dots&\mathbf{0}\\
 \mathbf{0}& uu^* &\dots&\mathbf{0}\\
 \vdots & \ddots&\cdots&\mathbf{0}\\
 \mathbf{0}& \mathbf{0}&\dots& uu^*
\end{bmatrix}, $$} i.e., $rank(\I_r^* u u^*\I_r)=1$ and $rank(\B(\omega))\leq n$. Let $\B(\omega)=\U(\omega) \Lambda(\omega)\U^*(\omega)$ be the eigen value decomposition of $\B(\omega),$ where $\Lambda=diag(\lambda_1,\dots,\lambda_n)$. Consequently, omitting $\omega$ from the notations,
\begin{align*}
\mbE\left[ \exp\left({tu^*\Q u}\right) \right]&=\mbE\left[ \exp\left({\frac{t}{n}\left[ \w^*\B\w - \mbE\left\{ \w^*\B\w \right\} \right] }\right) \right]\\ 
&=\mbE\left[ \exp\left({\frac{t}{n}\left[ \w^*\U \Lambda \U^*\w - \mbE\left\{ \w^*\U \Lambda \U^*\w \right\} \right] }\right) \right]\\ 
&\stackrel{(a)}{=}\mbE\left[ \exp\left({\frac{t}{n}\left[ \w^*\Lambda \w - \mbE\left\{ \w^* \Lambda \w \right\} \right] }\right) \right]\\
&=\mbE\left[ \exp\left({\frac{t}{n} \sum_{i=1}^{n} \lambda_i \left[w_i^2 - \mbE\left\{ w_i^2\right\} \right] }\right) \right]\\
&=\prod_{i=1}^{n} \mbE\left[ \exp\left({\frac{t\lambda_i }{n} \left[w_i^2 - \mbE\left\{ w_i^2\right\} \right] }\right) \right],
\end{align*}
where $(a)$ follows because $\w$ is invariant under unitary transformations \cite{cui2019covariance}.}
Let $\varepsilon \in \{+1,-1\}$ be a uniform random variable independent of $\w$. Similar to \eqref{recall1}, we can now apply the Rademacher random variable trick. 
\begin{align*}
 \mbE_{w_i}\left[ \exp\left(\lambda \left[w_i^2 - \mbE\left\{ w_i^2\right\} \right]\right) \right] &\leq \mbE_{w_i,\varepsilon}\left[ \exp\left(2\lambda \varepsilon w_i^2\right) \right]\\
 &=\sum_{k=0}^\infty \frac{\left(2\lambda\right)^{2k}}{(2k)!} \mathbb{E} \left[ w_i^{4k}\right]\\
 &\leq \sum_{k=0}^\infty \frac{\left(2\lambda\right)^{2k}}{2k!} \frac{(4k)!}{(2k)!2^{2k}}\\
 &\leq \sum_{k=0}^\infty {\left(2\lambda\right)^{2k}} = \frac{1}{1-4\lambda^2} \leq \exp(8 \lambda^2),
\end{align*}
for every $|\lambda|<3/8$. Thus, (with the substitution $\lambda=t\lambda_i/n$ and the upperbound $\lambda_i\leq \|\B\|$)
\begin{align*}
 \mbE\left[ \exp\left({tu^*\Q u}\right) \right]&= \prod_{i=1}^{n} \mbE\left[ \exp\left({\frac{t\lambda_i }{n} \left[w_i^2 - \mbE\left\{ w_i^2\right\} \right] }\right) \right]\\
 & \leq \prod_{i=1}^{n} \exp\left(\frac{8t^2\lambda_i^2}{n^2}\right)\\
 &=\exp\left(\frac{8t^2}{n^2}\sum_{i=1}^{n}\lambda_i^2\right)\\
 &\leq \exp\left(\frac{8t^2}{n}\|\cC\|^2\right), ~\forall~ |t|\leq \frac{3n}{8\|\cC\|},
\end{align*}
where we have used $\|B\|\leq\|\cC\|$ in the final equality. Now, combining this with the $\delta-$cover argument, 
\begin{align*}
 \mbE\left[ e^{t \|Q\|} \right] 
 &\leq \sum _{j=1}^m\mbE\left[ e^{2t(v^j)^*Qv^j} \right]+\mbE\left[ e^{-2t(v^j)^*Qv^j} \right]\\
 &\leq 2m \exp \left( \frac{32\|\cC\|^2t^2 }{n}\right)\\
 &\leq \exp \left( \frac{32\|\cC\|^2t^2}{n} + 6p\right), ~\forall |t| \leq \frac{3n}{16\|\cC\|}.
\end{align*}
Finally, applying Chernoff bound, $\mP\left( \left\|\Q\right\|\geq t \right) \leq \exp \left(-\frac{t^2n}{128 \|\cC\|^2 }+6p \right)$.
\subsubsection{Tight upper bound for $\|\cC\|$}
An explicit expression for $\cC$ is given as follows:
\begin{align*}
 \cC^{rs}(\omega):=\mathbb{E}\left\{\x^r(\omega)[\x^s(\omega)]^*\right\} &=\frac{1}{N} \sum_{\ell=0}^{N-1} \sum_{k=0}^{N-1} \mbE\left\{\breve{x}((r-1)N+\ell)[\breve{x}((s-1)N+k)]^T\right\} e^{-i\omega (\ell-k)}\\
 &=\frac{1}{N} \sum_{\ell=0}^{N-1} \sum_{k=0}^{N-1} R_{\brx}((r-s)N+\ell-k ) e^{-i\omega (\ell-k)}\\
 &\stackrel{}{=}\frac{1}{N} \sum_{\tau=-N+1}^{N-1} (N-|\tau|)R_{\brx}((r-s)N+\tau ) e^{-i\omega \tau}\\
 &{=} \sum_{\tau=-N+1}^{N-1} \left(1-\frac{|\tau|}{N}\right)R_{\brx}((r-s)N+\tau ) e^{-i\omega \tau}\\
 &=\sum_{\tau=-N+1}^{N-1} \left(1-\frac{|\tau|}{N}\right) e^{-i\omega \tau}R_{\brx}((r-s)N+\tau ).
 \end{align*}
Let $\alpha_\tau~=~ e^{-i\omega \tau}\left(1-\frac{|\tau|}{N}\right)$. Then 
\footnotesize
\begin{align*}
\cC&=\sum_{\tau=-N+1}^{N-1} \alpha_\tau \begin{bmatrix}
 R_{\brx}(\tau ) & R_{\brx}(-N+\tau )&\dots&R_{\brx}((1-n)N+\tau )\\
 R_{\brx}(N+\tau ) & R_{\brx}(\tau )&\dots&R_{\brx}((2-n)N+\tau )\\
 \vdots& \vdots&\ddots\\
 R_{\brx}((n-1)N+\tau ) & R_{\brx}((n-2)N+\tau )&\dots&R_{\brx}(\tau )
 \end{bmatrix}.
\end{align*}
\normalsize
Notice that $\breve{g}(\tau):=1-|\tau|/N$ is a triangle function, and the Fourier transform of $\breve{g}(\tau)$, $g(\omega)$ has the property that $|g(\omega)|\leq 1$, $\forall \omega \in \Omega$. Let $ \mathbf{u} \in \mC^{np}$ with $ \|\mathbf{u}\|_2\leq 1$ and $u(i) \in \mC^p$ be the $i$th p-dimensional subvector in $u$. Further, $\|u(\omega)\|^2_2=\left\|\sum_{k=1}^n e^{-i\omega k} u(k)\right\|_2^2 \leq \sum_{k=1}^n |e^{-i\omega k}| \left\|u(k)\right\|_2^2 =\left\|\mathbf{u}\right\|_2^2 \leq 1$. Then
\begin{align*}
 |\mathbf{u}^*\cC \mathbf{u}| &=\left|\sum_{i,j=1}^nu^*(i) \sum_{\tau=-N+1}^{N-1} \alpha_\tau R_{\brx}((i-j)N+\tau )u(j)\right|\\
 &=\left|\sum_{\tau=-N+1}^{N-1} e^{-i\omega \tau}\breve{g}(\tau)\sum_{i,j=-n}^n u^*(i) R_{\brx}\left(\left((i-j)+\frac{\tau}{N}\right)N \right) u(j)\right|, \because u(i):=0, \forall i\leq 0\\
 &=\left|\sum_{\tau=-N+1}^{N-1} e^{-i\omega \tau}\breve{g}(\tau)(v^* \star \y \star u) \left(\frac{\tau}{N}\right)\right|, \ \because v(i)=u({-i}), \forall i, \text{ and } \y(\tau)=R_{\brx}\left(N\tau\right)\\
 &\stackrel{(a)}{=}\left|\frac{1}{2\pi } \int_\Omega g(\nu-\omega)v^*(N\omega) \y(N \omega) u(N\omega) \mathrm{d}\omega \right|\\
&\leq \frac{1}{2\pi } \int_\Omega \left|v^*(N\omega) \y(N \omega)u(N\omega) \right| \mathrm{d}\omega \ \ ~(\because|g(\nu-\omega)|\leq 1)\\
 &\leq \frac{ M}{2\pi} \int_\Omega |v^*(N\omega) u(N\omega) | \mathrm{d}\omega\ ~(\text{by spectral norm definition and} \|\Phi_\x\|\leq M)\\ 
&\leq \frac{M}{2\pi} \int_\Omega \|v^*(N\omega)\|_2 \|u(N\omega) \|_2 \mathrm{d}\omega \\ 
&\leq M ~(\because\|u(\omega)\|_2\leq 1)
\end{align*} 

Thus, $\mP\left( \left\|\Q\right\|\geq t \right) \leq \exp \left(-\frac{t^2n}{128 M^2 }+6p \right)$, similar to the restart and record case.

\section{Proof of Lemma 4.4}
\label{app:lem:f_concentration_bound}
Notice that $\|A\x\|_2 \leq \|A\|\|\x\|_2$ for every matrix $A$ and vector $\x$. Applying this identity with $\x=[1,0\dots,0]$, $\|\Phi_{Ci}\|_2 \leq \|\Phi_{AA}\|$, where $A=[k,\ C]$. Then, applying CBS inequality for complex vectors, $|x^*Ay| \leq \|x\|_2\|Ay\|_2 \leq \|x\|_2\|A\|\|y\|_2$, the error can be upper bounded as
\begin{align*}
 | f(i,C,\omega)-\hatt{f}(i,C,\omega) |&=|(\Phi_{ii}- \Phi_{iC} \Phi_{CC}^{-1}\Phi_{Ci}) -(\hatt{\Phi}_{ii}- \hatt{\Phi}_{iC} \hatt{\Phi}_{CC}^{-1}\hatt{\Phi}_{Ci})|\\
 &=|(\Phi_{ii}-\hatt{\Phi}_{ii}) + (\hatt{\Phi}_{iC} \hatt{\Phi}_{CC}^{-1}\hatt{\Phi}_{Ci}- \Phi_{iC} \Phi_{CC}^{-1}\Phi_{Ci})|\\
 &\leq |\Phi_{ii}-\hatt{\Phi}_{ii}| + |\hatt{\Phi}_{iC} (\hatt{\Phi}_{CC}^{-1}- \Phi_{CC}^{-1})\hatt{\Phi}_{Ci})|\\
 & \hspace{1cm}+ |(\hatt{\Phi}_{iC}- \Phi_{iC}) \Phi_{CC}^{-1}\hatt{\Phi}_{Ci})|+|{\Phi}_{iC} {\Phi}_{CC}^{-1}(\hatt{\Phi}_{Ci}- \Phi_{Ci})|\\
 &\leq |\Phi_{ii}-\hatt{\Phi}_{ii}| + \|\hatt{\Phi}_{iC}\|_2 \|\hatt{\Phi}_{CC}^{-1}- \Phi_{CC}^{-1}\|\|\hatt{\Phi}_{Ci}\|_2 \\
 & \hspace{1cm}+ \|\hatt{\Phi}_{iC}- \Phi_{iC}\|_2 \|\Phi_{CC}^{-1}\|\|\hatt{\Phi}_{Ci}\|_2+\|{\Phi}_{iC}\|_2 \|{\Phi}_{CC}^{-1}\|\|\hatt{\Phi}_{Ci}- \Phi_{Ci}\|_2\\
 &\stackrel{}{\leq} |\Phi_{ii}-\hatt{\Phi}_{ii}| + \|\hatt{\Phi}_{CC}^{-1}- \Phi_{CC}^{-1}\|\|\hatt{\Phi}_{Ci}\|_2^2 \\
 & \hspace{1cm} +M\|\hatt{\Phi}_{iC}- \Phi_{iC}\|_2 \|\hatt{\Phi}_{Ci}\|_2+M^2\|\hatt{\Phi}_{Ci}- \Phi_{Ci}\|_2\\
 &\leq |\Phi_{ii}-\hatt{\Phi}_{ii}| + \|\hatt{\Phi}_{CC}^{-1}- \Phi_{CC}^{-1}\|\|\hatt{\Phi}_{Ci}-{\Phi}_{Ci}\|_2^2 + \|\hatt{\Phi}_{CC}^{-1}- \Phi_{CC}^{-1}\|\|{\Phi}_{Ci}\|_2^2\\
 & \hspace{1cm} +M\|\hatt{\Phi}_{iC}- \Phi_{iC}\|_2 \left(\|\hatt{\Phi}_{Ci}-\Phi_{Ci}\|_2+\|{\Phi}_{Ci}\|_2\right)+M^2\|\hatt{\Phi}_{Ci}- \Phi_{Ci}\|_2\\
 &\leq |\Phi_{ii}-\hatt{\Phi}_{ii}| + \|\hatt{\Phi}_{CC}^{-1}- \Phi_{CC}^{-1}\|\|\hatt{\Phi}_{Ci}-{\Phi}_{Ci}\|_2^2 + \|\hatt{\Phi}_{CC}^{-1}- \Phi_{CC}^{-1}\|M^2\\
 & \hspace{1cm} +M\|\hatt{\Phi}_{iC}- \Phi_{iC}\|_2 \left(\|\hatt{\Phi}_{Ci}-\Phi_{Ci}\|_2+M\right)+M^2\|\hatt{\Phi}_{Ci}- \Phi_{Ci}\|_2\\
 &= |\Phi_{ii}-\hatt{\Phi}_{ii}| + \|\hatt{\Phi}_{CC}^{-1}- \Phi_{CC}^{-1}\|\|\hatt{\Phi}_{Ci}-{\Phi}_{Ci}\|_2^2 + \|\hatt{\Phi}_{CC}^{-1}- \Phi_{CC}^{-1}\|M^2\\
 & \hspace{1cm} +M\|\hatt{\Phi}_{Ci}-\Phi_{Ci}\|_2^2+M^2\|\hatt{\Phi}_{Ci}-\Phi_{Ci}\|_2+M^2\|\hatt{\Phi}_{Ci}- \Phi_{Ci}\|_2\\&\leq |\Phi_{ii}-\hatt{\Phi}_{ii}| + \|\hatt{\Phi}_{CC}^{-1}- \Phi_{CC}^{-1}\|\|\hatt{\Phi}_{Ci}-{\Phi}_{Ci}\|_2^2 + M^2\|\hatt{\Phi}_{CC}^{-1}- \Phi_{CC}^{-1}\|\\
 \nonumber
 & \hspace{1cm} +M\|\hatt{\Phi}_{Ci}-\Phi_{Ci}\|_2^2+2M^2\|\hatt{\Phi}_{Ci}- \Phi_{Ci}\|_2.
\end{align*}
The above expression can be bounded above if we can bound the three errors, $\|\hatt{\Phi}_{ii}- \Phi_{ii}\|= \epsilon_i$, $\|\hatt{\Phi}_{AA}- \Phi_{AA}\|=\epsilon_A$, and $\|\hatt{\Phi}_{CC}^{-1}- \Phi_{CC}^{-1}\|=\epsilon_{Cinv}$. Simplifying the above expression,
\begin{align*}
 | f(i,C,\omega)-\hatt{f}(i,C,\omega)| &\leq |\Phi_{ii}-\hatt{\Phi}_{ii}| + \|\hatt{\Phi}_{CC}^{-1}- \Phi_{CC}^{-1}\|\|\hatt{\Phi}_{Ci}-{\Phi}_{Ci}\|_2^2 + M^2\|\hatt{\Phi}_{CC}^{-1}- \Phi_{CC}^{-1}\|\\
 \nonumber
 & \hspace{1cm} +M\|\hatt{\Phi}_{Ci}-\Phi_{Ci}\|_2^2+2M^2\|\hatt{\Phi}_{Ci}- \Phi_{Ci}\|_2\\
 &\leq \epsilon_i+\epsilon_{Cinv} \epsilon_A^2+2M^2\epsilon_{Cinv}+M\epsilon_A^2+2M^2\epsilon_A\\
 & \leq \epsilon_i+\epsilon_{Cinv} (\epsilon_A^2+2M^2)+3M^2\epsilon_A\\\
 & \leq \epsilon_i+3M^2\epsilon_{Cinv} +3M^2\epsilon_A
\end{align*}
Pick $\epsilon_i=3M^2\epsilon_{Cinv} =3M^2\epsilon_A =\epsilon/3$. 
Then $| f(i,C,\omega)-\hatt{f}(i,C,\omega)|<\epsilon$. From Section 5.8 in \cite{matrix_analysis},
\begin{align*}
\|\Phi_{CC}-\widehat{\Phi}_{CC}\| \label{eq:inverse_ineq}
&\leq \|\Phi_{CC}\|\|\Phi_{CC}^{-1}\|^{-1}\| \widehat{\Phi}_{CC}^{-1}-\Phi_{CC}^{-1}\| \frac{M^2 }{1-M^2 \frac{\|\widehat{\Phi}_{CC}^{-1}-\Phi^{-1}_{CC}\|}{\|\Phi^{-1}_{CC}\|}},\\
&\leq \frac{M^4 \|\widehat{\Phi}_{CC}^{-1}-\Phi_{CC}^{-1}\| }{1-M \|\widehat{\Phi}_{CC}^{-1}-\Phi_{CC}^{-1}\|} \leq\epsilon \implies \|\widehat{\Phi}_{CC}^{-1}-\Phi_{CC}^{-1}\|\leq\frac{\epsilon}{M^4+M\epsilon}\leq \frac{\epsilon}{M^4}.
\end{align*} Therefore, to guarantee that $\|\widehat{\Phi}_{CC}^{-1}-\Phi_{CC}^{-1}\|<\epsilon$, it is sufficient to guarantee that $\|\widehat{\Phi}_{CC}-\Phi_{CC}\|<\epsilon$ since $M\geq 1$. Rewriting Corollary 4.3, \begin{align}
 \mP\left(|\Phi_{ii}-\hatt{\Phi}_{ii}| \geq \epsilon \right) &\leq e^{ -\frac{\epsilon^2n}{128M^2}+6},\\
 \mP\left(\|\Phi_{AA}-\hatt{\Phi}_{AA}\| > \epsilon \right) &\leq e^{ -\frac{\epsilon^2n}{128M^2}+6(q+1)}, \text{ and }\\
 \mP\left(\|\Phi_{CC}-\hatt{\Phi}_{CC}\| >\epsilon\right) &\leq e^{ -\frac{\epsilon^2n}{128M^2}+6q}, \ \ \ \ \forall \epsilon\geq 0.
\end{align}

Plugging these bounds in the above expressions gives the concentration upper bound
\begin{align*}
 \mP\left( | f(i,C,\omega)-\hatt{f}(i,C,\omega)\geq \epsilon \right) &\leq \mP\left( |\Phi_{ii}-\hatt{\Phi}_{ii}| \geq \epsilon/3 \right)+\mP\left( \|\hatt{\Phi}_{CC}^{-1}- \Phi_{CC}^{-1}\| \geq \epsilon/{(9M^2)} \right)\\
 & \hspace{1cm}+\mP\left( \|\hatt{\Phi}_{AA}- \Phi_{AA}\| \geq \epsilon/{(9M^2)} \right)\\
 &\leq \mP\left( |\Phi_{ii}-\hatt{\Phi}_{ii}| \geq \epsilon/3 \right)+\mP\left( \|\hatt{\Phi}_{CC}- \Phi_{CC}\| \geq \epsilon M^2/{(9)} \right)\\
 & \hspace{1cm}+\mP\left( \|\hatt{\Phi}_{AA}- \Phi_{AA}\| \geq \epsilon/{(9M^2)} \right)\\
 &\leq e^{\left( -\frac{\epsilon^2n}{1152M^2}+6\right)}+e^{\left( -\frac{\epsilon^2M^2n}{10368 }+6q\right)}+e^{\left( -\frac{\epsilon^2n}{10368 M^{6}}+6(q+1)\right)}\\
 &\leq c_0e^{\left( -\frac{\epsilon^2n}{10368 M^{6}}+6(q+1)\right)}. 
\end{align*}
\hfill\qed

\section{Proof of Lemma 4.5}
Applying the bound in Lemma 4.4,
\begin{align*}
 \mP\left(G(\omega) \neq \hat{G}(\omega)\right) &= \mP\left( \bigcup_{k \in V, C \subseteq V\setminus \{k\}, |C| \leq q} \left\{ |f(i,C,\omega)-\hatt{f}(i,C,\omega)| > \epsilon\right\} \right) \\
 &\stackrel{(a)}{\leq} \sum_{k \in V, C \subseteq V\setminus \{k\}, |C| \leq q}\mP\left( \left\{ |f(i,C,\omega)-\hatt{f}(i,C,\omega)| > \epsilon\right\} \right) \\
 &\stackrel{(b)}{\leq} p \left[ {(p-1)\choose 1}+ \dots + {(p-1)\choose q}\right] c_0e^{\left( -\frac{\epsilon^2n}{10368 M^{6}}+6(q+1)\right)}\\
 &\stackrel{(c)}{\lesssim}c_1p \times q \times (p/q)^q e^{\left( -\frac{\epsilon^2n}{10368 M^{6}}+6(q+1)\right)}\\
 &\approx \exp\left(\log p +\log q + q\log \left(p/q\right)+\left( -\frac{\epsilon^2n}{10368 M^{6}}+6(q+1)\right)\right) \\
 &\approx \exp\left(\log p +\log q + q\log \left(p/q\right)+6q-\frac{\epsilon^2n}{M^{6}}\right) \\
 &\stackrel{}{\lesssim} \exp\left( q\log \left(p/q\right)-\frac{\epsilon^2n}{M^{6}}\right)<\delta,
\end{align*}
{where $(a)$ follows by union bound, $(b)$ follows since $|V|=p$ and there are ${p-1}\choose{k}$ number of combinations with $|C|=k$. $(c)$ follows by applying Stirling's approximation, ${n \choose k} \leq (ne/k)^k$.
}
Thus, $n \gtrsim \frac{M^6\left(q\log(p/q)-\log \delta\right)}{\epsilon^2} $. By selecting the threshold $\gamma$ in the algorithm appropriately, we can get a sample complexity $n \gtrsim \frac{M^6q\log(p/q)}{\Delta^2} $.

\section{Lower bound: Proof of Theorem 5.1}
\label{app:lowerbound}
The proof is based on Generalized Fano's inequality.
\begin{lem}[Generalized Fano's method] \cite{gao2022optimal}
Consider a class of observational distribution $\cF$ and a subclass $\cF'=\{F_1,\dots,F_r\} \subseteq \cF$ with $r$ distributions and the estimators $\widehat\theta$. Then
 \begin{align*}
 \inf_{\widehat{\theta}} \max_{F \in \cF} \mbE\{ \1(\theta(F)\neq \hatt{\theta})\} \geq \frac{\alpha_r}{2} \left( 1- \frac{n\beta_r+ \log 2}{\log r}\right),
 \end{align*}
 where $n$ is the number of samples, 
 \begin{align*}
 \alpha_r&:=\max_{k\neq j} \1(\theta(F_k)\neq \theta(F_j)),\\
 \beta_r&:=\max_{k\neq j} KL(F_k||F_j),
 \end{align*}
with $KL(P||Q):=\mathbb{E}_{P}\left[ \log\frac{P}{Q} \right]=\mathbb{E}_{P}\left[ \log{P}\right]-\mathbb{E}_{P}\left[ \log{Q} \right]$ being the KL divergence. 
\end{lem}

\begin{cor} Consider subclass of graphs $\cG'=\{G_1,\dots,G_r\} \subseteq \cG_{p,q}$, and let $\H^i$ be the distribution corresponding to a distinct $G_i \in \cG'$. Then, any estimator $\hatt{G}:=\bigcup\limits_{\omega \in \Omega}\hatt{G}(\omega)$ of $G_i$ is $\delta$ unreliable,
\label{cor:lowerbound}
\begin{align*}
{\inf_{\widehat{G}} }\sup_{G_i \in \cG'} \mP\{ (G(\H^i)\neq \hatt{G}\} \geq \delta, \end{align*}
 if 
 \begin{align*}
 n \leq \frac{(1-2\delta) \log r-\log 2}{\beta_r}
 \end{align*}
\end{cor}

Therefore, building a lower bound involves finding a subclass that has 1) small $\beta_r$ and 2) large $r$. First, we can find an upper bound of $r$ by upper bounding the number of directed graphs possible with at most $q$ parents. Overall, $p^2$ number of possible positions are there and at most $pq$ many edges. The number of possible ways to choose $k$ edges is ${p^2} \choose{k}$.
Thus $r=\sum \limits_{k=1}^{pq} {p^2 \choose k} \leq pq{p^2 \choose pq} \lesssim pq (p/q)^{pq}$.
Therefore, $\log r \lesssim \log (pq) + pq\log(p/q)$ $\lesssim pq\log(p/q)$
Similarly, it can be shown that $\log r \gtrsim pq \log (p/q)$ \cite{gao2022optimal}.

Consider the LDS (3), with $\breve{e}(k) \sim N(0,\sigma_k \I)$, i.i.d. across time. Then $\e(\omega) \sim N\left(0, \Phi_{\e}(\omega)\right)$ and $\x(\omega) \sim N(0, \Phi_\x)$, where $\Phi_{\e}(\omega)=\sum_{k\in \mathbb{Z}} \sigma_k \I$ and $\Phi_\x(\omega)=(I-\mathbf{H}(\omega))^{-1} \Phi_\e(\omega)((I-\mathbf{H}(\omega))^{-1})^*$.

\textbf{Ensemble A:} Consider all possible DAGs in $\cG'$ with i.i.d. Gaussian exogenous distribution such that $\Phi_\e(\omega)$ exists. For the two distributions, $F_k$ and $F_j$ such that for any $\omega \in \Omega$, $F_k(\omega) \sim \calN(\mathbf{0},{\Phi}^{(k)}(\omega))$ and $F_j(\omega) \sim \calN(\mathbf{0},{\Phi}^{(j)}(\omega))$,
\begin{align*}
 KL(F_k(\omega)||F_j(\omega))&=\frac{1}{2} \left( \mbE_{F_j(\omega)} [\x^*(\omega)[\Phi^{(k)}(\omega)]^{-1}\x(\omega)]-\mbE_{F_j} [\x^*(\omega)[\Phi^{(j)}(\omega)]^{-1}\x(\omega)] \right)\\
 &=\frac{1}{2} \left( \mbE_{F_j(\omega)} [tr(\x(\omega)[\Phi^{(k)}(\omega)]^{-1}\x^*(\omega))]-p \right)\\
 &=\frac{1}{2} \left( tr([\Phi^{(k)}(\omega)]^{-1}\Phi^{(j)}(\omega))-p \right)\\
 &\leq \frac{1}{2} \left( \sqrt{\|[\Phi^{(k)}(\omega)]^{-1}\|_F^2\|\Phi^{(j)}(\omega)\|_F^2}-p \right)\\
 &\leq \frac{1}{2} \left( pM^2-p \right) \leq (M^2-1)p
\end{align*}
Therefore one of the lower bounds is 

\begin{align*}
\inf_{\widehat{G}} \sup_{G \in \cG'} \mP\{ (G \neq \hatt{G})\} \geq \delta, \end{align*}
 if 
 \begin{align*}
 n &\leq \frac{(1-2\delta) pq \log (p/q))-\log 2}{(M^2-1)p}\\
 &\lesssim \frac{q \log(p/q)}{M^2-1}.
 \end{align*}

\textbf{Ensemble B:}
Here, we consider graphs in $\cH_{p,q}(\beta,\sigma,M)$ (recall Definition 2.4) with a single edge $u \xrightarrow{} v$ with $H_{vu}(\omega)=\beta$ for every $\omega \in \Omega$, i.e. constant matrix. For LDS with i.i.d. Gaussian noise with PSD matrix $\Phi_\x$ that satisfies this condition, $\H$ is such that $H_{vu} \neq 0$ and $H_{ij}=0$ otherwise. Here, the total number of graphs, $r=2{p \choose 2}=(p^2-p) \approx p^2$

Notice that $[\Phi_\x^{-1}]_{ij}=(\I_{ij}-H_{ij}-H_{ji}^*+ \sum_{k=1}^p H_{ki}^*H_{kj})/\sigma$. Thus (ignoring $\omega$), \\
$[\Phi_\x^{-1}]_{uv}=\frac{-H_{vu}^*+ \sum_{k=1}^p H_{ku}^*H_{kv}}{\sigma}=\frac{-H_{vu}^*}{\sigma}=-\beta/\sigma$ and $[\Phi_\x^{-1}]_{ij}=0$ if $i,j\neq u,v$. Then, 
\begin{align*}
\x^*\Phi_\x^{-1}\x&=\sum_{ij} x^*_{i}[\Phi_\x^{-1}]_{ij} x_j\\
&= \frac{1}{\sigma} \left[ \sum_{i=1}^p |x_i|^2 (1+\sum_k |H_{ki}|^2) + x_u^*[\Phi_\x^{-1}]_{uv} x_v + x_v^*[\Phi_\x^{-1}]_{vu} x_u\right]\\
&= \frac{1}{\sigma} \left[ \sum_{i\neq u} |x_i|^2 + (1+|H_{vu}|^2)|x_u|^2 - 2\beta\Re\{x_u^* x_v\} \right]\\
&=\frac{1}{\sigma} \left[ \sum_{i} |x_i|^2 + \beta^2 |x_u|^2 - 2\beta\Re\{x_u^* x_v\} \right]\\
&=\frac{1}{\sigma} \left[ \sum_{i\neq v} |x_i|^2 +|x_v- \beta x_u|^2\right]
\end{align*}
Therefore,
\begin{align*}
 KL(F^{uv}||F^{jk})&=\mbE_{F^{uv}} \left[ \log F^{uv} -\log F^{jk}\right] \\
 &=\frac{1}{2\sigma} \mbE_{F^{uv}}\left[ |x_v|^2 +|x_v- \beta x_u|^2- |x_k|^2 -|x_k- \beta x_j|^2\right]\\
 &=\frac{1}{2\sigma} \left[ \beta^2\sigma+ \mbE_{F^{uv}}\left( \beta^2 |x_j|^2 - 2\beta\Re\{x_j^* x_k\} \right)\right]
\end{align*}
Considering all the cases of $(u,v)$ vs $(j,k)$ it can be shown that $KL(F^{uv}||F^{jk}) \leq \beta^2+ \beta^4/2$ \cite{gao2022optimal}. Thus, $n \gtrsim \frac{\log p}{\beta^2+ \beta^4/2}$ gives the second lower bound. The lower bound follows by combining ensembles $A$ and $B$. \hfill \qed
\end{document}